\documentclass{article}
\usepackage{PRIMEarxiv}
\usepackage[dvipsnames]{xcolor}

\usepackage{booktabs}
\usepackage{graphicx}
\usepackage{multirow}
\usepackage{color, colortbl, xcolor}
\usepackage{soul}
\usepackage{multirow}
\usepackage{algorithm}
\usepackage{algorithmicx}
\usepackage{algpseudocode}
\usepackage{mathtools}

\usepackage{etoolbox}
\usepackage{bbm}
\usepackage{listings} 
\usepackage{amssymb}
\usepackage{amsbsy}
\usepackage{pifont} 
\usepackage[caption=false]{subfig}

\usepackage{tikz}
\usetikzlibrary{shapes.geometric}
\usepackage{pgfplots}
\usepackage{pgfplotstable}
\pgfplotsset{compat=newest}
\usepackage[normalem]{ulem}

\definecolor{2nd}{gray}{0.8}

\newcolumntype{g}{>{\columncolor{Gray}}c}

\newcolumntype{^}{>{\currentrowstyle}}

\algdef{SE}[SUBALG]{Indent}{EndIndent}{}{\algorithmicend\ }%
\algtext*{Indent}
\algtext*{EndIndent}

\newcolumntype{Y}{>{\centering\arraybackslash}X}
\newcolumntype{P}[1]{>{\centering\arraybackslash}p{#1}}

\algnewcommand{\nComment}[1]{\Statex \Comment{#1}}
\definecolor{LightCyan}{rgb}{0.88,1,1}

\definecolor{mypurple}{RGB}{153, 0, 153}
\definecolor{mygray}{RGB}{128, 128, 128}
\definecolor{mygreen}{RGB}{0, 153, 0}

\definecolor{mycyan}{RGB}{64, 128, 128}
\definecolor{mypink}{RGB}{255, 182, 193}
\definecolor{myred}{RGB}{165,42,42}
\definecolor{myyellow}{RGB}{255, 191, 0}

\definecolor{tab_red}{rgb}{0.71, 0.11, 0.0}
\definecolor{tab_green}{rgb}{0.11, 0.71, 0.0}

\definecolor{car_yellow}{RGB}{255,255,153}
\definecolor{dog_purple}{RGB}{204,153,255}
\definecolor{bg_white}{RGB}{255,255,255}

\usetikzlibrary{shadows}
\usetikzlibrary{plotmarks}
\usetikzlibrary{shapes}

\newcommand{\thickhline}{%
	\noalign {\ifnum 0=`}\fi \hrule height 1pt
	\futurelet \reserved@a \@xhline
}
\makeatletter
\global\let\oriCT@@do@color\CT@@do@color

\usepackage{amsmath,amsfonts}
\usepackage[utf8]{inputenc} 
\usepackage[T1]{fontenc}    
\usepackage[breaklinks, colorlinks=true, citecolor=MidnightBlue, linkcolor=BrickRed, urlcolor=Blue]{hyperref}
\usepackage[capitalize,noabbrev]{cleveref}
\usepackage{url}            
\usepackage{booktabs}       
\usepackage{nicefrac}       
\usepackage{microtype}      
\usepackage{lipsum}
\usepackage{fancyhdr}       
\usepackage{graphicx}       
\graphicspath{{media/}}     
\usepackage{textcomp}
\usepackage{stfloats}
\usepackage{url}
\usepackage[numbers, sort&compress]{natbib}
\usepackage{enumitem}
\usepackage{dirtytalk}
\definecolor{mpl_blue}{HTML}{1f77b4}
\usepackage{authblk}

\title{Learning to Complement with Multiple Humans}
\author[1]{Zheng Zhang}
\author[1]{Cuong Nguyen}
\author[2]{Thanh-Toan Do}
\author[1]{Kevin Wells}
\author[1]{Gustavo Carneiro}
\affil[1]{Centre for Vision, Speech and Signal Processing, University of Surrey, United Kingdom}
\affil[2]{Department of Data Science and AI, Monash University, Australia}

\begin{document}
\maketitle

\begin{abstract}
Real-world image classification tasks tend to be complex, where expert labellers are sometimes unsure about the classes present in the images, leading to the issue of learning with noisy labels (LNL). The ill-posedness of the LNL task requires the adoption of strong assumptions or the use of multiple noisy labels per training image, resulting in accurate models that work well in isolation but fail to optimise human-AI collaborative classification (HAI-CC). Unlike such LNL methods, HAI-CC aims to leverage the synergies between human expertise and AI capabilities but requires clean training labels, limiting its real-world applicability. This paper addresses this gap by introducing the innovative \underline{Le}arning to \underline{Co}mplement with \underline{M}ultiple \underline{H}umans (LECOMH) approach. LECOMH is designed to learn from noisy labels without depending on clean labels, simultaneously maximising collaborative accuracy while minimising the cost of human collaboration, measured by the number of human expert annotations required per image. Additionally, new benchmarks featuring multiple noisy labels for both training and testing are proposed to evaluate HAI-CC methods. Through quantitative comparisons on these benchmarks, LECOMH consistently outperforms competitive HAI-CC approaches, human labellers, multi-rater learning, and noisy-label learning methods across various datasets, offering a promising solution for addressing real-world image classification challenges.
\end{abstract}


\section{Introduction}

\citet{gao2022learning} When dealing with real-world image classification problems, it is common that labellers often encounter difficulties in accurately labelling images~\citep{carneiro_machine_2024}. This can occur for various reasons, such as the difficulty of the problem (e.g., medical diagnosis~\citep{chen2023bomd} or fine-grained classification~\citep{wei2023fine}) or due to the labeller's lack of experience~\citep{kamar2012combining}. 
In this paper, we primarily focus on the challenges posed by the complexity of the classification task, assuming that labellers are experts in classifying dataset images.

\begin{figure*}[tbp]
    \centering
    \subfloat[]{\label{fig:noisy_label}\includegraphics[width=.325\linewidth]{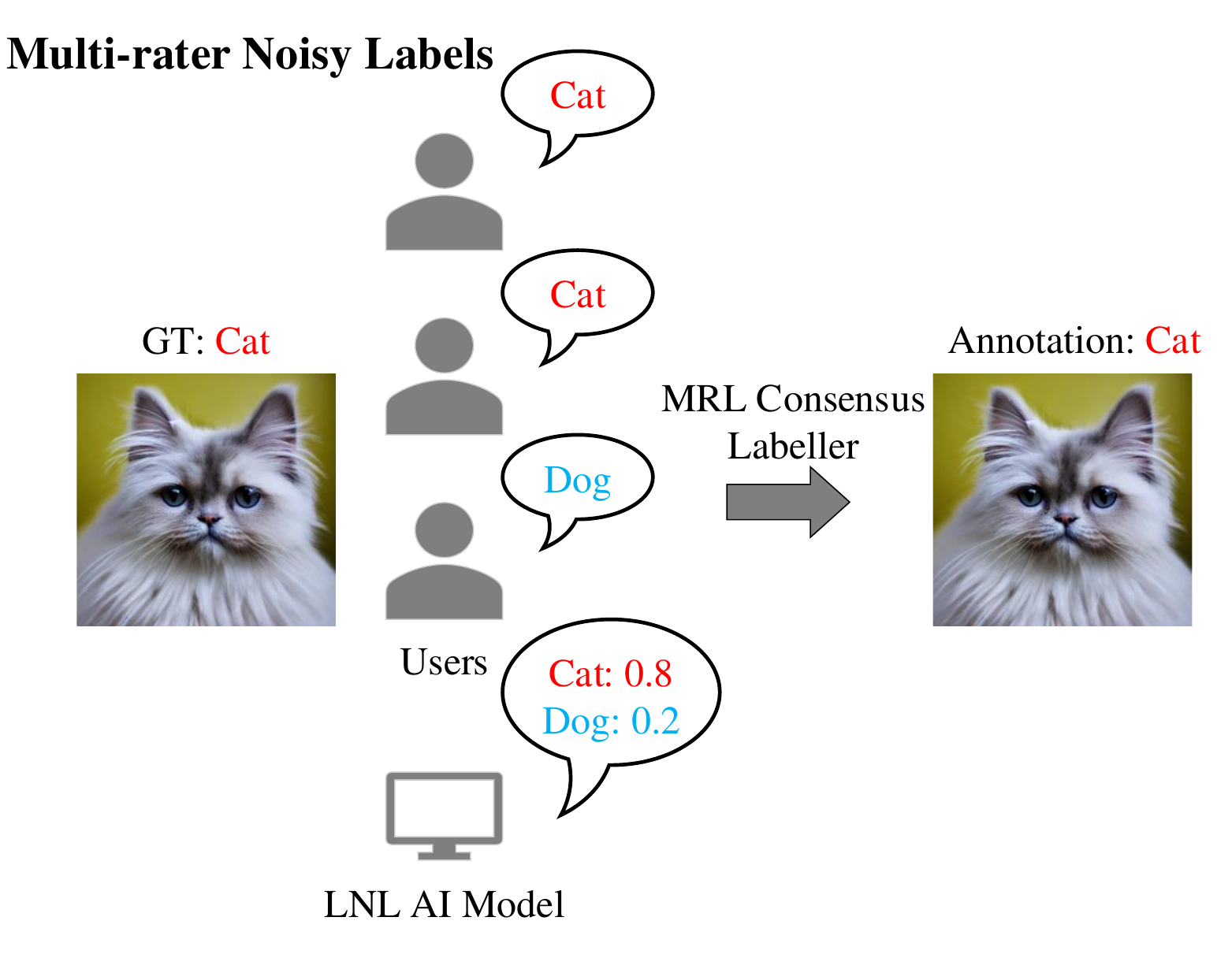}}\hfill
    \subfloat[]{\label{fig:network}\includegraphics[width=.625\linewidth]{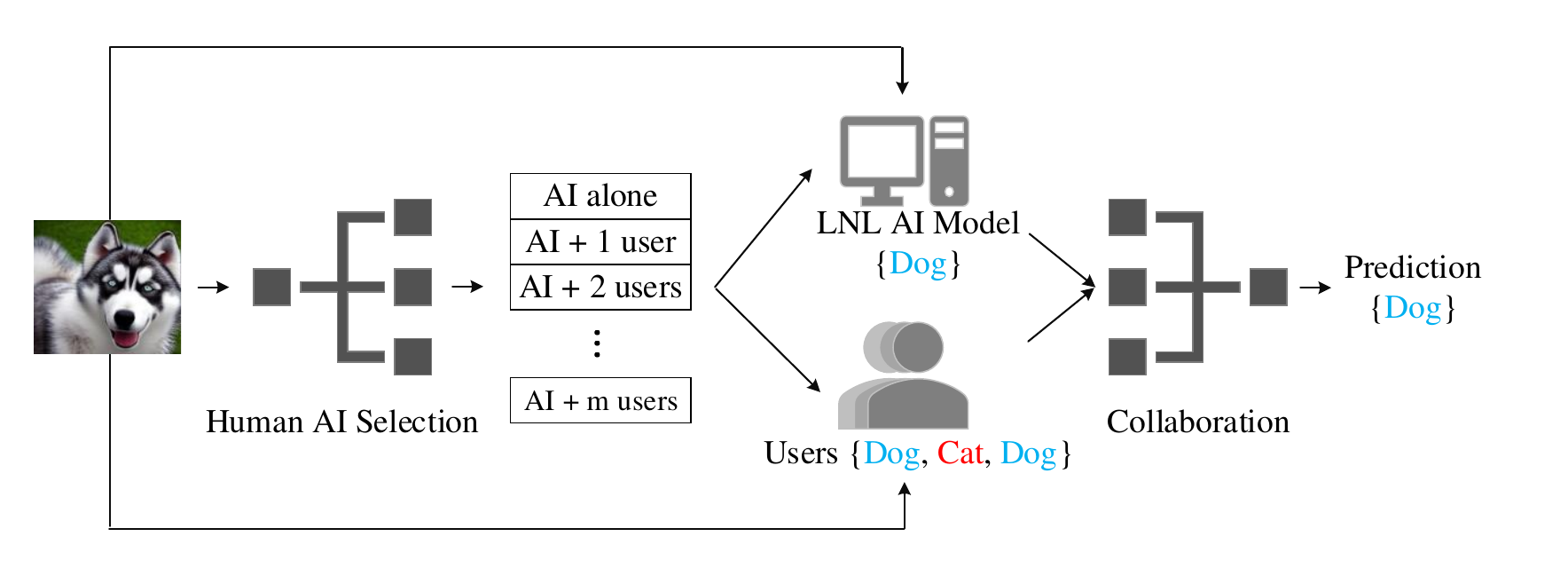}}\\
    \caption{LECOMH is the first human-AI collaborative classification (HAI-CC) method that learns exclusively from multiple noisy labels and collaborates with multiple experts.
    Its primary objective is to optimise HAI-CC accuracy while concurrently minimising collaboration costs, measured by the number of human expert annotations required for image classification. 
    To enable the learning from multiple noisy labels, we first train an AI model using learning with noisy label (LNL) techniques, followed by a multi-rater learning (MRL) to produce a consensus label that is then used as the ground truth label for training the two stages of HAI-CC. The first stage is the \emph{Human-AI Selection Module} that estimates the number of human predictions needed for efficient and accurate human-AI collaborative classification, and the second stage is the \emph{Collaboration Module} that produces the final prediction.}
    \label{fig:motivation}
    \end{figure*}

Two separate research communities have addressed this challenge by making different assumptions and proposing quite distinct solutions. 
The learning with noisy-label (LNL) research community focuses on mitigating the presence of noise in the labels with sophisticated training methods~\citep{song2022learning,carneiro_machine_2024,ji2021learning} that aim to maximise model accuracy during testing when they operate in isolation.
Given the ill-posedness of the LNL problem~\citep{liu2023identifiability}, current solutions either need to impose strong assumptions (e.g., clean-label training samples tend to have smaller losses than noisy-label samples) or they rely on multiple noisy labels per training image to build multi-rater learning (MRL) methods~\citep{ji2021learning}.
On the other hand, the human-AI collaborative classification (HAI-CC) community~\citep{dafoe2021-cooperative} focuses on the development of methods that assume the presence of clean and multiple noisy training labels to exploit the complementary performance of human experts and AI to produce a collaborative approach that has higher accuracy than both the expert’s and the AI’s accuracy.
Apart from the need of clean labels, another limitation of HAI-CC methods is that they  collaborate only with single users, limiting real-world deployment.
There are notable HAI-CC exceptions~\citep{ijcai2022-344,multil2d} that can learn to complement with or defer to multiple experts, but their reliance on clean-label samples, absence of collaboration cost optimisation~\citep{ijcai2022-344}, and lack of human-AI ensemble classification~\citep{multil2d}
restrict their applicability in real-world scenarios. 
As a result, it is possible to notice a remarkable research gap, where LNL methods can handle noisy labels in training, but they do not collaborate with users during testing, while HAI-CC methods collaborate with users during testing, but they rely on clean labels for training and rarely collaborate with multiple experts. 
    
    
    This paper addresses the research gap exposed above with the innovative \underline{Le}arning to \underline{Co}mplement with \underline{M}ultiple \underline{H}umans (LECOMH) approach and the introduction of new HAI-CC benchmarks.
    LECOMH, shown in \cref{fig:motivation}, is designed to learn from multiple noisy labels per sample to maximise the HAI-CC accuracy and minimise the multiple user collaboration costs, measured by the number of human expert annotations in the collaborative classification of a test image.
    The proposed benchmarks assess HAI-CC methods with datasets containing multiple noisy labels in training and testing. 
    Overall, the key contributions of the paper are:
    \begin{itemize}
        \item the first HAI-CC method, referred to as LECOMH, that can be trained exclusively from multiple noisy labels per training image to maximise the collaborative classification accuracy of teams of AI and multiple  experts, while minimising the collaboration costs, measured by the number of human experts used in HAI-CC, and
        \item new benchmarks to assess HAI-CC methods on classification problems containing multiple noisy labels in the training and testing sets, paving the way for a more comprehensive performance evaluation of real-world applications.
    \end{itemize}
    The empirical evaluation shows that LECOMH consistently demonstrates superior performance than state-of-the-art (SOTA) HAI-CC methods~\citep{whoshould_mozannar23,ijcai2022-344,multil2d} in the newly-introduced benchmarks with higher accuracy for equivalent collaboration costs. 
    Furthermore, LECOMH is the only HAI-CC method in our experiments that outperforms expert labellers and isolated LNL methods across all datasets.
    
    The rest of this work is organised as follows: \cref{sec:related_work} presents a brief review of previous studies on noisy-label learning, multi-rater learning and human-AI collaborative classification. \cref{sec:methods} describes the proposed Learning to Complement with Multiple Humans approach. \cref{sec:benchmark} introduces new benchmarks to assess human-AI collaborative classifiers with multi-rater noisy-label datasets, including new benchmarks in CIFAR-10, Chaoyang, and NIH datasets. \cref{sec:experiments} presents and analyses the empirical evaluation. Finally, \cref{sec:conclusion} summarises and conclude this study.

\section{Related Work}
\label{sec:related_work}

LECOMH is a new method that jointly addresses the challenges of learning with multiple noisy labels and human-AI collaboration. Thus, in this section, we review relevant studies in learning with noisy labels, multi-rater learning, and human-AI collaborative classification.

\subsection{Learning with Noisy Labels (LNL)}
\label{sec:LNL_review}

LNL is a challenging problem that has received increasing attention by the machine learning community~\citep{song2022learning,carneiro_machine_2024}.
Since learning with noisy labels is an ill-posed problem, it necessitates the imposition of specific constraints to facilitate the discovery of a viable solution.
One common constraint is the small-loss hypothesis~\citep{li2020dividemix,MentorNet,arazo2019unsupervised}, which posits that clean-label training samples incur smaller losses than noisy-label samples. 
Another constraint, known as clusterability~\citep{zhu2021clusterability}, assumes that a training sample and its two nearest neighbours share the same clean label.

The development of these constraints allowed the proposal of a vast number of LNL methods, which include: robust loss functions~\citep{zhang2018generalized,ghosh2017robust},
co-teaching~\citep{MentorNet, han2018co},
label cleaning~\citep{yuan2018iterative, jaehwan2019photometric},
semi-supervised learning (SSL)~\citep{li2020dividemix, ortego2021multi},
iterative label correction~\citep{label_clean, arazo2019unsupervised},
meta-learning ~\citep{L2W, Distill_noise, FSR, Famus}, and
graphical modelling~\citep{garg2023instance}.
Among these, SSL represents a dominant technique used in LNL~\citep{li2020dividemix}. Graphical models also show accurate results for some specific LNL problems (e.g., instance-dependent noise)~\citep{garg2023instance}.

Recent research has studied alternative ways to imposing the constraints mentioned above for solving LNL problems.
A particularly interesting alternative approach is based on a training algorithm that exploits  multiple noisy labels per training sample~\citep{liu2023identifiability}. 
An important distinction needs to be made at this stage, which is related to the expertise of the labellers who produce the multiple noisy labels per training sample. 
When labellers are not not experts, exhibiting a broad range of accuracy and reliability~\citep{rodrigues2017learning, dawid1979maximum, whitehill2009whose, raykar2010learning, rodrigues2018deep, herde2023multi}, the learning method is usually referred to as crowd-sourcing that aims to quantify the quality of labellers with the goal of optimising either data assignment or labeller reward.
On the other hand, when labellers are experts, their labels tend to be reliable even if small inter-user variations are present, and the learning method explored in this case is the multi-rater learning (MRL).  Since in this paper, we assume that labellers are experts, we focus on MRL methods that will be  explained in the next section.

Although much research has been developed to address the LNL problem, none of the methods above collaborate with users during
testing, even though such collaboration has the potential to improve the LNL results~\citep{rastogi2023taxonomy}.


\subsection{Multi-rater Learning (MRL)}
\label{sec:MRL_review}

MRL methods aim to assign consensus labels from the multiple (potentially divergent or conflicting) noisy labels produced by expert users. 
MRL follows two modelling approaches, namely: inter-annotator and intra-annotator. 
In the inter-annotator approach, learning is performed to characterise inter-observer variabilities~\citep{raykar2009supervised, guan2018said, mirikharaji2021d, ji2021learning}, while in the intra-annotator approach, learning aims to estimate annotator-specific variabilities (e.g., through confusion matrices)~\citep{khetan2017learning, tanno2019learning, wu2022learning, cao2023learning}. 
A joint learning method combining both the inter- and intra-annotator approaches was proposed to integrate the strengths of both approaches~\citep{wu2022learning}. 
One drawback in those studies is the assumption of sample-independence, potentially deviating from real-world applications where annotation error might depend on both samples and labellers. Such a challenge motivates the study by~\citet{gao2022learning} to learn a sample-dependent model. 
In another study, UnionNet~\citep{wei2022deep} has been developed to integrate labelling information from all annotators, leveraging this collective input to better coordinate responses across multiple contributors. 
CrowdAR~\citep{cao2023learning} focuses on predicting the reliability of annotations to directly assess the quality of crowd-sourced data, which is used as a soft annotation to produce a consensus training label. Furthermore, ADMoE~\citep{zhao2023admoe} adopts a Mixture of Experts (MoE) architecture to foster specialised and scalable learning from multiple noisy sources, particularly targeting anomaly detection. CROWDLAB~\citep{goh2022CROWDLAB} is a state-of-the-art (SOTA) MRL method that produces consensus labels using a combination of multiple noisy labels and the predictions by an external classifier. 

It is important to mention that the MRL methods discussed earlier typically do not explicitly incorporate SOTA LNL methods, a consideration we aimed to address in our approach. 
Furthermore, similarly to the LNL methods presented in Sec.~\ref{sec:LNL_review}, MRL approaches have not been designed to collaborate with users during
testing. Such collaboration has the potential to improve the accuracy shown by MRL methods~\citep{rastogi2023taxonomy}, as we demonstrate in our paper.

\subsection{Human-AI Collaborative Classification (HAI-CC)}

In machine learning, the vast majority of AI systems have been optimised in isolation, without considering the implications of human-AI collaborative classification~\citep{rosenfeld2018totally, serre2019deep,kamar2012combining}.
Scientifically, the isolated development of AI systems is correct, but in practice, the influence of AI decisions on humans is unpredictable and represents a critical point to study.
Recently, \citet{chiou2023trusting} studied how AI decisions influence human experts, reaching the conclusion that the trustworthiness of AI depends on both model confidence~\citep{lu2021human, yin2019understanding} and explainability~\citep{shin2021effects, weitz2019you}. 
Nevertheless, this is a two-way lane, and while it is important to consider how AI influences humans, we must also consider that humans can affect AI decision, which is the main study topic of HAI-CC~\citep{bansal2021most, agarwal2023combining, vodrahalli2022humans, humanIntheloop, complement_wilder}. 
In general, when classifying an image, there are three options to be considered by HAI-CC systems:
\begin{enumerate}[label=\Roman*]
    \item) \textit{AI predicts alone};
    \item) \textit{AI defers decisions to human experts};
    \item) \textit{AI makes a joint decision with human experts}.
\end{enumerate}
Below, we explain the main HAI-CC approaches being studied, relating them to these three options. 


\subsubsection{Learning to Defer (L2D)}

L2D methods focus on HAI-CC options (I) and (II) above, meaning that the decision relies on a \say{single} prediction made by the AI model or the human expert. L2D aims to learn a classifier and a rejector to decide when a human expert prediction should replace the AI prediction~\citep{madras2018predict, keswani2021towards, narasimhan2022post, mao2023two}. 
This \emph{rejection learning}~\citep{cortes2016learning} approach was generalised by considering the human expertise in the decision-making process~\citep{madras2018predict}.
Further investigation into L2D methods then concentrated on the development of different surrogate loss functions that are consistent with the Bayes-optimal classifier obtained in the case of 0-1 loss~\citep{narasimhan2022post, charoenphakdee2021classification, raghu2019algorithmic, okati2021differentiable, mozannar2020consistent, verma2022calibrated, whoshould_mozannar23, charusaie2022sample, cao2024defense, straitouri2023improving, liumitigating, mozannar2022teaching}. 
Another L2D approach is the score-based triages~\citep{raghu2019algorithmic} which introduced two error prediction algorithms for human and machine errors to optimise the decision and reduce overall error. 
This was extended by the differentiable triages that utilise a deterministic threshold rule for triage decisions, where the threshold is derived from the differences in errors between the model and human decisions on individual instances~\citep{okati2021differentiable}.
One common limitation of L2D-based methods is the reliance on the single-expert setting, overlooking the more complex environments with the availability of multiple experts. 
Hence, recent research in L2D has shifted the focus to the multiple-expert setting~\citep{verma2022calibration, mao2023two, multil2d, keswani2021towards, babbar2022utility, mao2023principled, hemmer2023learning, tailor2024learning, leitao2022human}.
Despite such an extensive research, current L2D-based learning methods have not been designed to enable AI models and human experts to jointly produce a final classification. 
To address this gap, \say{learning to complement} methods have been introduced as explained below.

\subsubsection{Learning to Complement (L2C)}

L2C methods focus on HAI-CC options (I) and (III), aiming to optimise the collaboration between human experts and the AI model to maximise the expected utility of the human-AI decision~\citep{complement_wilder,steyvers2022bayesian,kerrigan2021combining,liu2023humans,bansal2021most,ijcai2022-344}. 
\citet{kerrigan2021combining} proposed to combine human and model predictions via confusion matrices and model calibration. \citet{steyvers2022bayesian} introduced a Bayesian framework for combining the predictions and different types of confidence scores from humans and machines, demonstrating that a hybrid combination of human and machine predictions leads to better performance than combinations of human or machine predictions alone. Recently, \citet{liu2023humans} leverages perceptual differences between humans and AI to make a human-AI system outperform humans or AI alone, while \citet{ijcai2022-344} introduced a model featuring an ensemble prediction involving both AI and human predictions, yet it does not optimise the collaboration cost.

    
One common assumption of existing HAI-CC methods, including both L2D- and L2C-based approaches, is the availability of \emph{clean} labels. This is, however, impractical because real-world settings typically only contain multiple noisy labels per sample. Furthermore, although collaborative classification has been explored
\citep{ijcai2022-344}, the cost produced by such a collaboration is not taken into account. In contrast, our proposed LECOMH is designed to work with multiple noisy labels per sample in the training set by exploring LNL and MRL methods. Furthermore, LECOMH jointly maximises human-AI classification accuracy and minimises the collaboration cost, measured by the number of human experts annotations used to classify an image. 


\section{Learning to Complement with Multiple Humans (LECOMH)}
\label{sec:methods}
\begin{figure*}[t]
    \centering
    \includegraphics[width=0.8\linewidth]{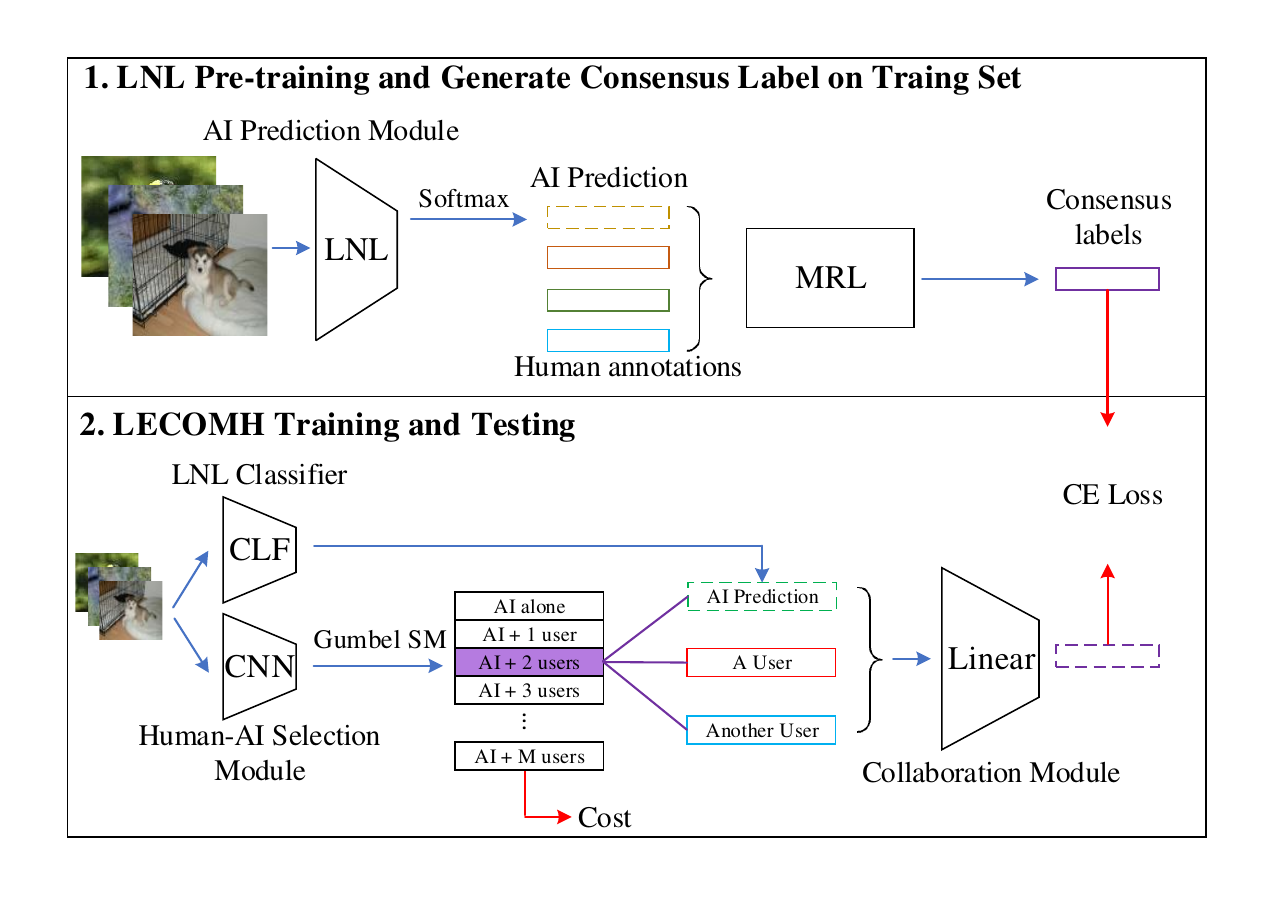}
    \caption{\if LECOMH involves a multi-step training process. It starts with pre-training the LNL AI model~\citep{wang2022promix,garg2023instance}, followed by training the MRL method, e.g. CROWDLAB~\citep{goh2022CROWDLAB}, to combine user labels with AI predictions, creating a consensus label. The LECOMH training process aims to maximise classification accuracy and minimise collaboration costs through iterative steps. These steps include: \fi
    The proposed LECOMH consists of two main steps: 1. \emph{(top)} estimate the consensus labels by exploiting a pre-trained LNL model coupled with an MRL module~\citep{goh2022CROWDLAB}; and 2. \emph{(bottom)} train an LNL classifier (CLF) and a human-AI selection module by minimising both the classification error and the collaboration cost. In particular, the training step involves:
    1) building the set of AI predictions and user labels, 2) training the Human-AI Selection Module to estimate the number of users to collaborate with the  AI classifier, and 3) training the Collaboration Module to produce a final classification using AI predictions and selected users' labels. Testing involves similar steps to generate the final prediction.
 }
	\label{fig:architecture}
\end{figure*}

    Let $\mathcal{D}=\{x_i,\mathcal{M}_i\}^{|\mathcal{D}|}_{i=1}$ be the noisy-label multi-rater training set, where $x_i\in\mathcal{X}\subset\mathbb{R}^{H\times W\times R}$ denotes an input image of size $H\times W$ and $R$ channels, 
    and $\mathcal{M}_i=\{ m_{i,j}\}_{j=1}^{M}$ denotes the $M$ experts' noisy annotations for the image $i$, with $m_{i,j} \in \mathcal{Y} \subseteq \{0,1\}^{|\mathcal{Y}|}$ being a one-hot label.   
    Our methodology, as layout in \cref{fig:network}, consists of: 1) an AI Model pre-trained with LNL techniques to enable the production of a training sample consensus label by the multi-rater learning approach CROWDLAB~\citep{goh2022CROWDLAB}, 2) a Human-AI Selection Module that predicts the collaboration format (i.e., AI alone, AI + 1 user, AI + 2 users, etc.), and 3) a Collaboration Module that aggregates the predictions selected by the  Human-AI Selection Module to produce a final prediction. 
    Note that our training and testing, as explained below, are designed to be unbiased to any specific labeller, so when we select AI + 1 user or AI + 2 users, the users are randomly selected from our pool of users.

\subsection{Training}

    LECOMH maximises classification accuracy and minimises collaboration costs in a human-AI collaborative classification setting, where cost is proportional to the number of users that are asked to provide labels.
    Our training has three phases (see \cref{fig:architecture}): 1) pre-training the LNL AI Model, 2) generating consensus labels for the training set using the multi-rater learning CROWDLAB method~\citep{goh2022CROWDLAB}, and 3) training of LECOMH's Human-AI Selection and Collaboration Modules. We provide more details as follows:

\subsubsection{LNL Pre-training and  Consensus Label Generation} 
    We use SOTA LNL techniques~\citep{wang2022promix,garg2023instance,zhu2021hard,liu2022nvum} to train the LNL AI model $f_\theta:\mathcal{X} \to \Delta^{|\mathcal{{Y}}|-1}$, where $\Delta^{|\mathcal{Y}|-1}$ denotes the $|\mathcal{Y}|$-dimensional probability simplex, and $\theta\in\Theta$ is the classifier's parameter. 
    This LNL training uses the training set, where the noisy label of image $x_i$ is randomly selected as one of the experts' annotations in $\mathcal{M}_i$.
    For the consensus label generation, we leverage the SOTA multi-rater learning method CROWDLAB~\citep{goh2022CROWDLAB} that takes the training images and experts' labels  $(x,\mathcal{M}) \in \mathcal{D}$, together with the AI classifier's predictions $\hat{y} = f_{\theta}(\mathbf{x})$ for each sample in $\mathcal{D}$ to produce a consensus label $\hat{y}^c \in \mathcal{Y}$ and a quality (or confidence) score $\alpha$. Formally, the formation of the consensus label dataset can be written as:
\begin{equation}
    \begin{aligned}[b]
        \mathcal{D}^c =\{(x_i,\hat{y}^c_i,\mathcal{M}_i) | (x_i,\mathcal{M}_i)\in\mathcal{D} 
        \wedge (\hat{y}_i,\alpha_i) = \mathsf{CROWDLAB}(x_i,f_{\theta}(x_i),\mathcal{M}_i) \wedge \alpha_i>0.5\},
    \end{aligned}
  \label{eq:consensus}
\end{equation}
    which is used by the LECOMH training, as explained below. 
    We use CROWDLAB for MRL because it can combine labels from annotators and the pre-trained LNL AI model to produce highly accurate consensus labels~\citep{goh2022CROWDLAB} for the subsequent LECOMH training.

\subsubsection{LECOMH training} 
    The proposed LECOMH comprises the Human-AI Selection Module and the Collaboration Module, as shown in \cref{fig:architecture}. 
    The Human-AI Selection Module, represented by $g_{\phi}:\mathcal{X} \to \Delta^{M}$, predicts a probability of having either an isolated AI prediction (1st dimension) or a combined prediction between AI and multiple users (remaining $M$ dimensions). In other words, the \(j\)-th index of \(g^{(j)}_{\phi}(x)\) represents selecting the AI model and \(j-1\) annotators.
    The Collaboration Module, represented by
    \(h_{\psi}: \left( \Delta^{|\mathcal{Y}| - 1} \right)^{M + 1} \to \Delta^{|\mathcal{Y}| - 1},\)
    takes the AI prediction in the first input, and the remaining user predictions selected by $g_{\phi}(.)$ to produce the final classification prediction \(\Tilde{y}_{i}\) defined as follows:
    \begin{equation}
        \Tilde{y}_{i} = h_{\psi} \left( \mathsf{p} \left(g_{\phi}(x_i),f_{\theta}(x_i), \mathsf{rand}(\mathcal{M}_i) \right) \right),
        \label{eq:final_label}
    \end{equation}
    where:
    \begin{equation}
    \begin{aligned}[b]
        & \mathsf{p} \left (g_{\phi}(x),f_{\theta}(x), \mathsf{rand}(\mathcal{M}) \right ) = 
        & \hspace{-0.5em}\begin{cases}
            \begin{bmatrix}
                f_{\theta}(x) & \mathbf{0}_{|\mathcal{Y}|} & \dots & \mathbf{0}_{|\mathcal{Y}|}
            \end{bmatrix}^{\top}
            & \text{if } \max_j g^{(j)}_{\phi}(x) = g_{\phi}^{(1)}(x) \\
            \begin{bmatrix}
                f_{\theta}(x) & m_{i,1} & \dots & \mathbf{0}_{|\mathcal{Y}|}
            \end{bmatrix}^{\top} & \text{if } \max_j g^{(j)}_{\phi}(x) = g_{\phi}^{(2)}(x) \\
            \dots & \\
            \begin{bmatrix}
                f_{\theta}(x) & m_{i,1} & \dots & m_{i,M}
            \end{bmatrix}^{\top} & \mbox{if } \max_j g^{(j)}_{\phi}(x) = g_{\phi}^{(M+1)}(x),
       \end{cases}
    \end{aligned}
    \label{eq:concatenate_predictions}
    \end{equation}
    with $g^{(j)}_{\phi}(.)$ denoting the $j$-{th} output from the Human-AI Selection Module and $\mathsf{rand}(\mathcal{M})$ representing a function that randomly selects the experts' annotations to avoid bias toward any specific experts' annotations.

    The Human-AI Selection Module and the Collaboration Module is trained by minimising the cross-entropy loss \(\ell(., .)\) between the consensus label \(y_{i}^{c}\) and the final prediction \(\Tilde{y}_{i}\), plus an additional term that regularises the cost as follows:
\begin{equation}
    \begin{aligned}[b]
        \if \phi^*, \psi^* = \arg \fi \min_{\phi,\psi} \frac{1}{|\mathcal{D}^c|} \sum_{(x_i,\hat{y}^c_i,\mathcal{M}_i) \in \mathcal{D}^c} & \ell \left( \hat{y}^c_i, \Tilde{y}_{i} \right) + \lambda \times \mathsf{cost}(g_{\phi}(x_i)),
    \end{aligned}
    \label{eq:loss_function}
\end{equation}
where \(\Tilde{y}_{i}\) is the consensus label of sample \(x_{i}\), defined in \cref{eq:final_label}, $\lambda$ is a hyper-parameter that weights the cost function, and
\begin{equation}
    \mathsf{cost}(g_{\phi}(x)) = \sum_{j=1}^{M+1}g^{(j)}_{\phi}(x) \times (j-1).
    \label{eq:cost}
\end{equation}
For the cost in \cref{eq:cost}, when the AI model is selected to predict alone, the selection module would output a probability such that $\max_j g^{(j)}_{\phi}(x) = g_{\phi}^{(1)}(x)$, resulting in $\mathsf{cost}(g_{\phi}(x)) = 0$. Thus, for the case that $\max_j g^{(j)}_{\phi}(x) = g_{\phi}^{(K)}(x)$ for $K\in[2,M]$, then $\mathsf{cost}(g_{\phi}(x)) \approx K-1$. In other words, the cost in \cref{eq:cost} represents the cost of one unit per expert's annotation.

As explained above and depicted in \cref{fig:architecture} \emph{(bottom)}, the human-AI training has two main steps: selecting the collaboration format (i.e., AI and the number of experts) and making the prediction through the collaboration module.
The selection of the collaboration format is estimated by sampling from the probability vector output of the selection module \(g_{\phi}(x)\). 
Naively sampling from such a categorical distribution is non-differentiable, prohibiting the training using stochastic gradient descent. 
To avoid that, we employ the concrete distribution~\citep{maddison2017the}, also known as Gumbel-softmax trick~\citep{jang2016categorical}, to approximate such sampling, making it trainable with SGD. 

\subsection{Testing}
    Testing starts from the LNL AI prediction, followed by the human-AI selection module prediction of the categorical distribution of the probability of the AI model running alone or collaborating with a set of $K \in [1,M]$ users, resulting in a cost of \(K\).
    After deciding on the number of users to collaborate, using Gumbel-softmax on $g_{\phi}(x)$, we randomly select testing users, and concatenate their predictions with the AI prediction to serve as input to the collaboration module, which outputs the final classification, following the procedure defined in \cref{eq:final_label,eq:concatenate_predictions}.

\section{Human-AI Collaborative Benchmarks}
\label{sec:benchmark}

\subsection{New CIFAR-10 Benchmarks}
    We introduce two new benchmarks with CIFAR-10~\citep{krizhevsky2009learning}, which has 50K training images and 10K testing images of size $32\times 32$. 
    The \textit{first benchmark relies on annotations produced by people} to produce CIFAR-10N~\citep{wei2021learning} for training and CIFAR-10H~\citep{peterson2019human} for testing.
    CIFAR-10N~\citep{wei2021learning} has three noisy labels for each CIFAR-10 training image, while CIFAR-10H~\citep{peterson2019human} provides approximately 51 noisy labels per CIFAR-10 testing image. 
    Due to the limitation of three labels per sample in CIFAR-10N, the testing process allows collaboration with at most three users randomly sampled from the pool of users in CIFAR-10H. 
    The \textit{second benchmark}, named multi-rater CIFAR10-IDN~\citep{xia2021sample}, \textit{is based on synthesised annotations} for training and testing with multi-rater instance-dependent noise. The label noise rates 0.2 and 0.5 for both training and testing sets, with three distinct noisy labels generated for each noise rate to simulate varying human predictions with similar error rates.

\subsection{New Chaoyang Benchmark} 
    The Chaoyang dataset has 6,160 colon slides represented as patches of size $512\times 512$~\citep{zhu2021hard}, where each patch has \textit{three noisy labels produced by real pathologists}. 
    Originally, the dataset had a training set with 4,021 patches for training and 2,139 patches for testing. The training patches had multi-rater noisy labels, while testing patches only contained a unanimous expert agreement on a single label. To create a new benchmark, the dataset was restructured to ensure both training and testing sets contained multiple noisy labels. The entire dataset was reshuffled, resulting in a partition of 4,725 patches for training and 1,435 patches for testing. In this new partition, both sets have multi-rater noisy label patches. The training set comprises 862 patches with 2 out of 3 consensus labels and 3,862 patches with 3 out of 3 consensus labels. The testing set includes 449 patches with 2 out of 3 consensus labels and 986 patches with 3 out of 3 consensus labels. Importantly, patches from the same slide do not appear in both the training and testing sets.

\subsection{Multi-rater NIH Dataset} 

    The multi-rater NIH Chest X-ray dataset~\citep{majkowska2020chest,wang2017chestx} contains an average of 3 manual labels per image for four radiographic findings on 4,374 chest X-ray images~\citep{majkowska2020chest} from the ChestX-ray8 dataset~\citep{wang2017chestx}. We focus on the occurrence of the following clinically important findings: airspace opacity (NIH-AO) and nodule or mass (NIH-NM). The prevalence of NIH-AO and NIH-NM findings are close to 50\% and 14\%. We selected a total of 2,412 images in the validation set for training and 1,962 images in the testing set for testing. The prediction accuracy of the 3 users in the NIH-AO dataset is approximately 89\%, 94\%, 80\% in training and 89\%, 94\%, 80\% in testing, while in the NIH-NM dataset, the prediction accuracy of the 3 users are 92\%, 92\%, 93\% in training and 89\%, 90\%, 91\% in testing. 

\section{Experiments}
\label{sec:experiments}
\subsection{Implementation Details}
\subsubsection{Models used} 
\label{sec:models_used}
    All methods are implemented in PyTorch~\citep{paszke2019pytorch} and run on NVIDIA RTX A6000. For experiments performed on CIFAR-10N and CIFAR-10H datasets, we employ ProMix~\citep{wang2022promix} to pre-train two PreAct-ResNet-18 as the LNL AI models using the set of \emph{Rand1} annotations in CIFAR-10N~\citep{wei2021learning}. For the experiments of multi-rater learning performed on CIFAR10-IDN, we use InstanceGM~\citep{garg2023instance} to pre-train two PreAct-ResNet-18 as the LNL AI models. For the Chaoyang dataset, we follow the practice in NSHE~\citep{zhu2021hard} to pre-train two ResNet-34 using the set of \emph{label\_A} annotations. The network having the highest performance is selected for the LNL AI model. For the NIH datasets, we follow the NVUM model~\citep{liu2022nvum} by pre-training on the ChestXray dataset~\citep{wang2017chestx} and then fine-tuning on the airspace opacity and nodule and mass classification tasks. All the above models are selected according to their SOTA performance in the respective datasets. The same pre-trained backbones are also used for the Human-AI Selection Module. The Collaboration Module is designated as a two-layer multi-layer perceptron, where each hidden layer has 512 neutrons activated by the Rectified Linear Unit (ReLU) function.

    We also measure the performance of those pre-trained models as references. 
    The pre-trained ProMix on CIFAR-10N reaches 97.41\% accuracy on the CIFAR-10 test set. In the multi-rater setting relying on CIFAR-10 IDN, the pre-trained InstanceGM reaches an accuracy of 96.64\% and 95.90\% for the label noise rates at 0.2 and 0.5, respectively. The pre-trained NSHE achieves 82.44\% prediction accuracy on Chaoyang, while the pre-trained NVUM is 86.65\% and 87.41\% on airspace opacity and nodule or mass findings, respectively.

\subsubsection{Training and evaluation details}

For each dataset, the proposed human-AI system is trained for 200 epochs using SGD with a momentum of 0.9 and a weight decay of \(5 \times 10^{-4}\). The batch size used is 256 for CIFAR, 96 for Chaoyang and 32 for NIH. The initial learning rate is set at 0.05 and decayed through a cosine annealing. The temperature parameter of the Gumbel-softmax sampling is set at 5. Also, to ensure consistent data range between the LNL classifier and users' predictions, the LNL classifier predictions are normalised with a softmax activation 
before it is concatenated with the users' annotations. 
Given that we do not want to bias the performance of the system to any particular user, 
the users in each collaboration format (e.g., the \(m\) users in the format of AI model plus \(m\) users) are randomly selected during training and testing for LECOMH. 
For training, the ground truth labels are set as the consensus labels obtained via CROWDLAB. 
For testing, the ground truth label is either available from the dataset (e.g., for the CIFAR benchmarks) or from the consensus label obtained from majority voting (e.g., for Chaoyang and Multi-rater NIH benchmarks).
The evaluation is based on the prediction accuracy as a function of coverage evaluated on the test sets. 
Coverage denotes the percentage of examples classified by the AI model, with $100\%$ coverage representing a classification performed exclusively by the classifier, while $0\%$ coverage denoting a classification performed exclusively by the users.
To obtain different levels of coverage for LECOMH, we adjust the hyper-parameter $\lambda$ in \cref{eq:loss_function} during training, where the higher the hyper-parameter \(\lambda\), the more emphasis on the cost, and hence, the lower the coverage.
To report the mean and standard error of the system accuracy, each run is repeated for 5 trials with different random seeds.


\begin{figure*}[tbp]
    \centering
    \subfloat[CIFAR-10H.]{\label{fig:cifar10nh}\includegraphics[width=.475\linewidth]{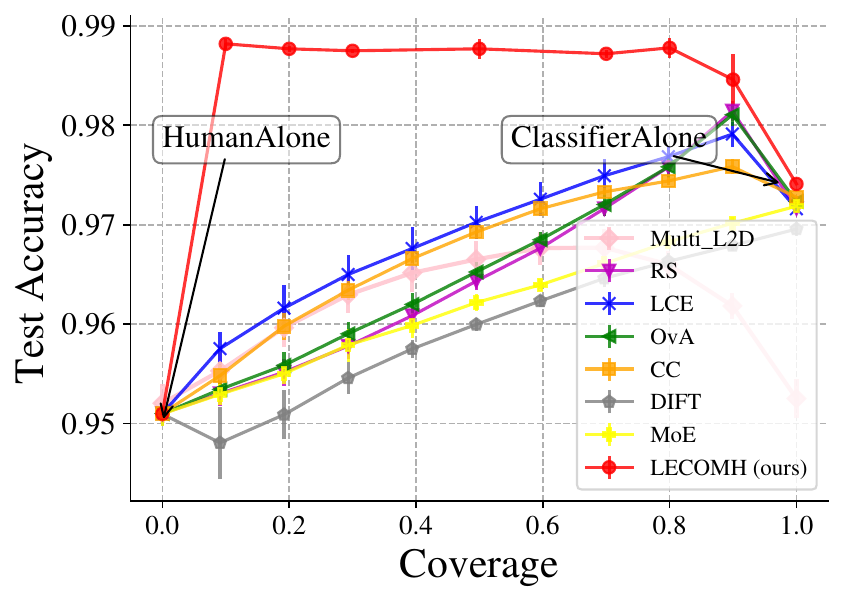}}\hfill
    \subfloat[IDN20.]{\label{fig:idn20}\includegraphics[width=.475\linewidth]{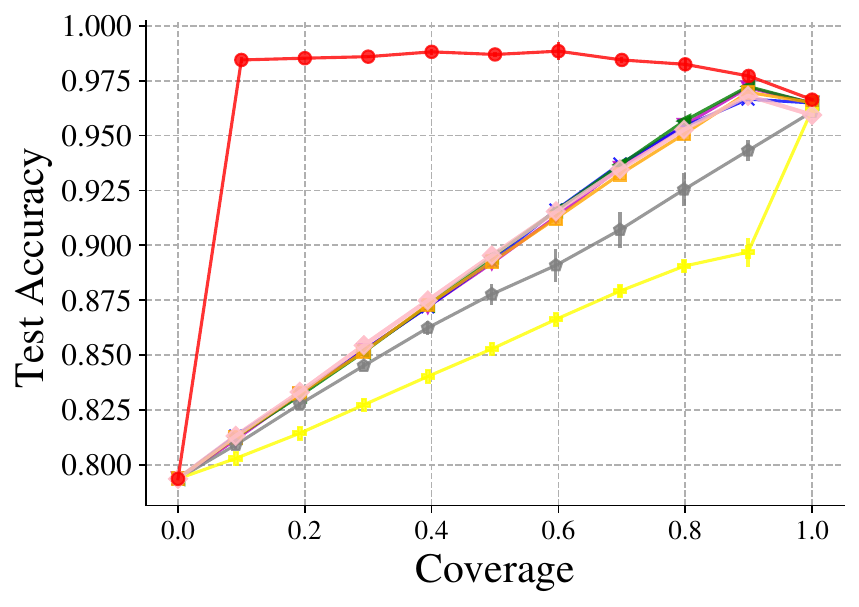}}\\
    \subfloat[IDN50.]{\label{fig:idn50}\includegraphics[width=.475\linewidth]{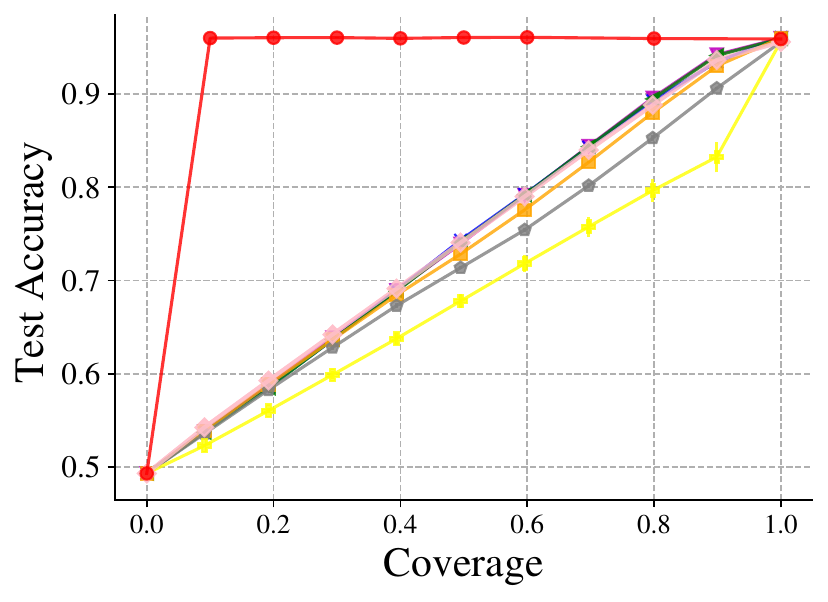}}\hfill
    \subfloat[Chaoyang.]{\label{fig:chaoyang}\includegraphics[width=.475\linewidth]{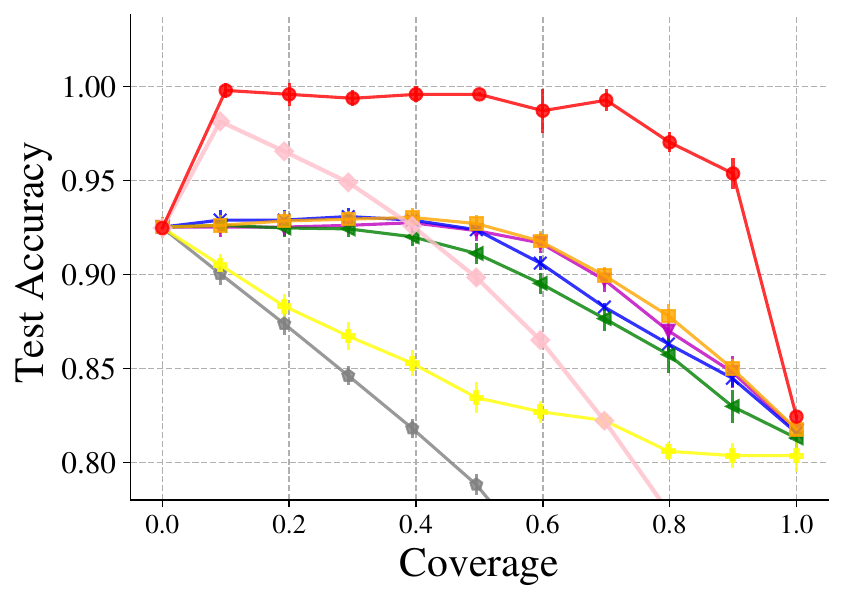}}\\
    \subfloat[NIH-AO.]{\label{fig:NIH-AO}\includegraphics[width=.475\linewidth]{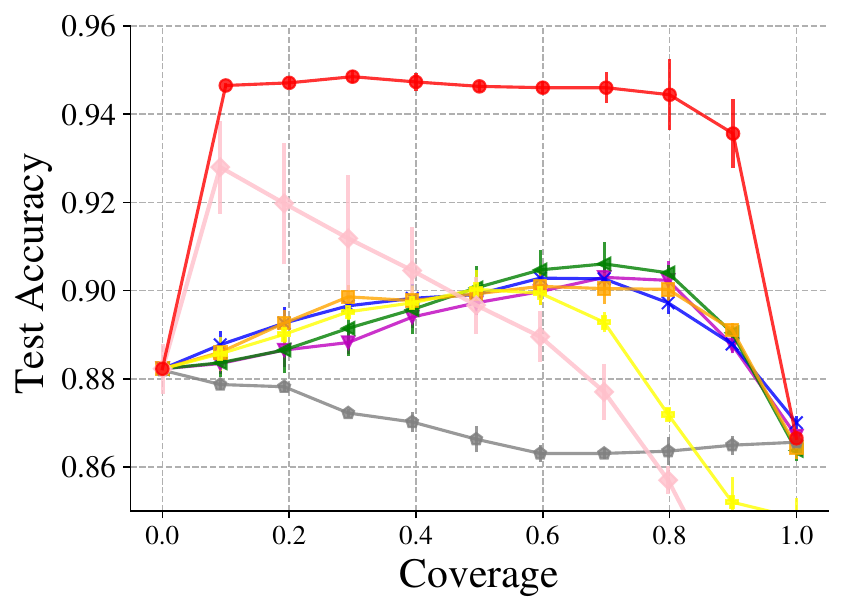}}\hfill
    \subfloat[NIH-NM.]{\label{fig:NIH-NM}\includegraphics[width=.475\linewidth]{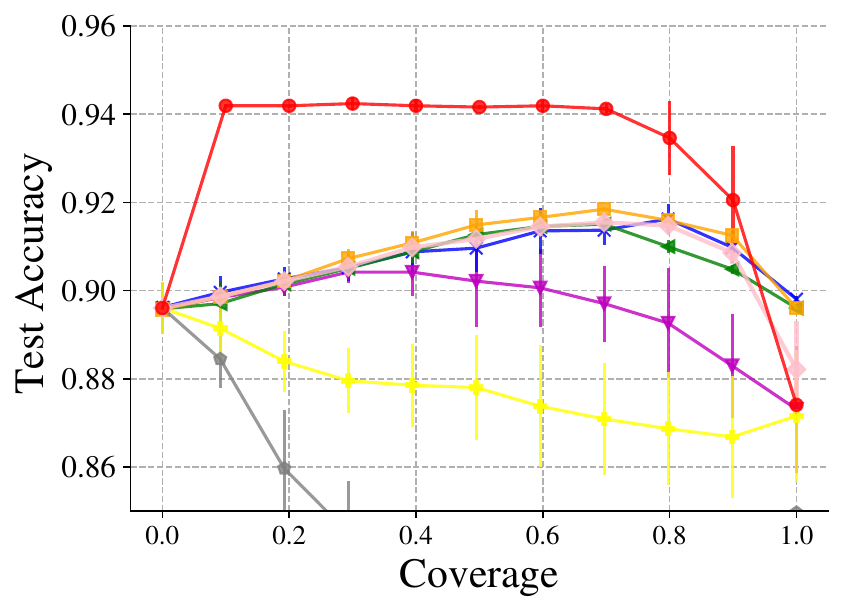}}\\
    \caption{Test accuracy vs. coverage of LECOMH (Ours) and competing SEHAI-CC~\citep{whoshould_mozannar23} and MEHAI-CC~\citep{ijcai2022-344,multil2d} methods. The SEHAI-CC methods are always pre-trained with LNL techniques, with the single user being simulated with aggregation (majority voting) from the pool of three annotators. Multi\_L2D can defer to one of many experts, so we select the label corresponding to the maximum probability of 3 users for each sample to draw the curve.}
    \label{fig:exp}
\end{figure*}

\subsubsection{Baselines}

We follow~\citep{whoshould_mozannar23} and evaluate LECOMH in the single expert human-AI collaborative classification (SEHAI-CC) setting. We also consider several SOTA methods, such as cross-entropy surrogate (CE)~\citep{mozannar2020consistent}, one-vs-all-based surrogate (OvA)~\citep{verma2022calibrated}, the confidence method (CC)~\citep{raghu2019algorithmic}, differentiable triage (DIFT)~\citep{okati2021differentiable}, mixture of experts (MoE)~\citep{madras2018predict}, and Realizable Surrogate (RS)~\citep{whoshould_mozannar23} as baselines. For a fair comparison, we randomly sample a single annotation for each image as a way to simulate a single expert from the human annotation pools to train those SEHAI-CC methods. 
To calculate the SEHAI-CC's collaboration coverage, we follow the procedure presented in~\citep{whoshould_mozannar23} by sorting the testing images based on their rejection scores and then adjusting the threshold of the quantile used for annotating these testing cases by users. 
We also compare LECOMH with methods that defer to multiple experts (MEHAI-CC), including classifier and expert team (CET)~\citep{ijcai2022-344}, and learning to defer to multiple experts (Multi\_L2D)~\citep{multil2d} in our setting.
For a fair comparison, all classification backbones for the \{SE,ME\}HAI-CC methods have the same architecture. All \{SE,ME\}HAI-CC methods rely on LNL pre-training because they provide better results for all cases. For all \{SE,ME\}HAI-CC methods, hyper-parameters are set as previously reported in~\citep{whoshould_mozannar23,ijcai2022-344,multil2d}. We also provide the results of SOTA LNL approaches, such as ProMix~\citep{wang2022promix} (CIFAR-10N), InstanceGM~\citep{garg2023instance} (CIFAR10-IDN), NSHE~\citep{zhu2021hard} (Chaoyang) and NVUM~\citep{liu2022nvum} (NIH datasets), and SOTA MRL approaches, such as majority voting, UnionNet~\citep{wei2022deep} and CROWDLAB~\citep{goh2022CROWDLAB}. Note that when considering CROWDLAB as an isolated baseline, its AI model is the pre-trained vanilla ResNet18 trained with early stopping, not the LNL models mentioned in \cref{sec:models_used}.

\begin{table*}[t]
    \centering
    \label{tab:quantitive_comparison}
    \scalebox{.9}{
    \begin{tabular}{l l c c c c c c}
        \toprule
        \bfseries Methods & \bfseries Type & \bfseries CIFAR-10H & \bfseries IDN20 & \bfseries IDN50 & \bfseries Chaoyang & \bfseries NIH-AO & \bfseries NIH-NM \\
        \midrule
        AI & LNL & 97.39 $\pm$ 0.16 & 96.64 $\pm$ 0.04 & 95.90 $\pm$ 0.25 & 82.44 $\pm$ 0.20 & 86.65 $\pm$ 0.35 & 87.41 $\pm$ 0.19\\
        Human* & Annotator & 95.10 & 79.36 & 49.30 & 92.47 & 88.23 & 89.60 \\
        \midrule
        RS & SEL2D & 96.65 $\pm$ 0.19 & 89.19 $\pm$ 0.10 & 73.92 $\pm$ 0.09 & 92.33 $\pm$ 0.57 & 89.72 $\pm$ 0.26 & 90.21 $\pm$ 1.10 \\
        MoE & SEL2D & 96.00 $\pm$ 0.07 & 85.28 $\pm$ 0.15 & 67.79 $\pm$ 0.78 & 83.45 $\pm$ 0.82 & 90.03 $\pm$ 0.43 & 87.80 $\pm$ 1.19  \\
        LCE & SEL2D & 96.43 $\pm$ 0.09 & 89.26 $\pm$ 0.08 & 74.32 $\pm$ 0.06 & 92.38 $\pm$ 0.33 & 89.90 $\pm$ 0.29 & 90.96 $\pm$ 0.54  \\
        OvA & SEL2D & 97.02 $\pm$ 0.16 & 89.47 $\pm$ 0.04 & 74.11 $\pm$ 0.06 & 91.12 $\pm$ 0.57 & 90.06 $\pm$ 0.49 & 91.27 $\pm$ 0.15  \\
        CC & SEL2D & 96.52 $\pm$ 0.10 & 89.25 $\pm$ 0.13 & 72.83 $\pm$ 0.11 & 92.71 $\pm$ 0.47 & 89.93 $\pm$ 0.31 & 91.49 $\pm$ 0.33  \\
        DIFT & SEL2D & 96.93 $\pm$ 0.06 & 87.77 $\pm$ 0.48 & 71.36 $\pm$ 0.17 & 78.82 $\pm$ 0.54 & 86.63 $\pm$ 0.30 & 83.64 $\pm$ 0.55  \\ \midrule
        Multi\_L2D & MEL2D & 96.22 $\pm$ 0.09 & 89.54 $\pm$ 0.15 & 74.05 $\pm$ 0.30 & 89.84 $\pm$ 0.39 & 89.67 $\pm$ 0.65 & 91.17 $\pm$ 0.19  \\ \midrule
        CET* & MEL2C & 97.76 $\pm$ 0.07 & 96.13 $\pm$ 0.23 & 95.18 $\pm$ 0.17 & 99.20 $\pm$ 0.80 & 94.14 $\pm$ 0.31 & 90.57 $\pm$ 0.10   \\ \midrule
        Majority Vote & MRL & 97.48 $\pm$ 0.00 & 92.48 $\pm$ 0.05 & 62.55 $\pm$ 0.44 & 99.58 $\pm$ 0.00 & 94.13 $\pm$ 0.00 & 94.14 $\pm$ 0.00    \\
        UnionNet & MRL & 93.34 $\pm$ 0.45 & 95.59 $\pm$ 0.11 & 93.54 $\pm$ 0.21 & 74.08 $\pm$ 0.92 & 85.94 $\pm$ 0.58 & 87.50 $\pm$ 0.52   \\
        CROWDLAB & MRL & 97.72 $\pm$ 0.02 & 92.40 $\pm$ 0.18 & 61.95 $\pm$ 0.31 & 99.58 $\pm$ 0.00 & 94.09 $\pm$ 0.00 & 92.61 $\pm$ 0.00 \\ \midrule
        \rowcolor{Gray!25} LECOMH & MEL2C & \textbf{98.77 $\pm$ 0.10} & \textbf{98.82 $\pm$ 0.15} & \textbf{96.05 $\pm$ 0.05} & \textbf{99.58 $\pm$ 0.42} & \textbf{94.63 $\pm$ 0.12} & \textbf{94.19 $\pm$ 0.05}  \\
        \bottomrule
    \end{tabular}}
    \caption{Quantitative comparison with the SOTA SEL2D~\citep{whoshould_mozannar23}, ME\{L2D,L2C\}~\citep{ijcai2022-344,multil2d} and MRL~\citep{goh2022CROWDLAB, wei2022deep} methods on the human-AI collabotative classification datasets at 50\% coverage. The SEL2D methods are always pre-trained with LNL techniques, with the single user being simulated with random selection from the pool of three annotators. 
    The ME\{L2D,L2C\} methods are also pre-trained with LNL techniques. The notation * means the results are under zero coverage (i.e., using the annotations from all human experts). The best result per benchmark is marked in bold.}
\end{table*}

\begin{table*}[t]
    \centering
    \label{tab:best_acc_comparison}
    \scalebox{.9}{
    \begin{tabular}{l l c c c c c c}
        \toprule
        \bfseries Methods & \bfseries Type & \bfseries CIFAR-10H & \bfseries IDN20 & \bfseries IDN50 & \bfseries Chaoyang & \bfseries NIH-AO & \bfseries NIH-NM \\ \midrule
        AI & LNL & 97.39 $\pm$ 0.16 & 96.64 $\pm$ 0.04 & 95.90 $\pm$ 0.25 & 82.44 $\pm$ 0.20 & 86.65 $\pm$ 0.35 & 87.41 $\pm$ 0.19 \\
        Human & Annotator & 95.10 & 79.36 & 49.30 & 92.47 & 88.23 & 89.60 \\ \midrule
        RS & SEL2D & 98.14 $\pm$ 0.10 & 97.18 $\pm$ 0.01 & 95.99 $\pm$ 0.04 & 92.75 $\pm$ 0.51 & 90.30 $\pm$ 0.21 & 90.42 $\pm$ 0.25 \\
        MoE & SEL2D & 97.19 $\pm$ 0.03 & 96.18 $\pm$ 0.12 & 95.64 $\pm$ 0.04 & 92.47 $\pm$ 0.54 & 90.28 $\pm$ 0.23 & 89.60 $\pm$ 0.59 \\
        LCE & SEL2D & 97.91 $\pm$ 0.13 & 96.67 $\pm$ 0.01 & 96.06 $\pm$ 0.02 & 93.08 $\pm$ 0.44 & 90.60 $\pm$ 0.50 & 91.62 $\pm$ 0.35 \\
        OvA & SEL2D & 98.11 $\pm$ 0.11 & 97.24 $\pm$ 0.03 & 95.95 $\pm$ 0.02 & 92.61 $\pm$ 0.52 & 90.10 $\pm$ 0.43 & 91.51 $\pm$ 0.06 \\
        CC & SEL2D & 97.58 $\pm$ 0.08 & 96.98 $\pm$ 0.12 & 96.02 $\pm$ 0.02 & 93.03 $\pm$ 0.53 & 88.19 $\pm$ 0.16 & 91.85 $\pm$ 0.04 \\
        DIFT & SEL2D & 96.96 $\pm$ 0.07 & 96.09 $\pm$ 0.21 & 95.58 $\pm$ 0.14 & 92.47 $\pm$ 0.54 & 90.03 $\pm$ 0.43 & 89.60 $\pm$ 0.59 \\ \midrule
        Multi\_L2D& MEL2D & 96.77 $\pm$ 0.07 & 96.82 $\pm$ 0.06 & 95.59 $\pm$ 0.04 & 98.14 $\pm$ 0.05 & 92.80 $\pm$ 1.05 & 91.55 $\pm$ 0.19 \\ \midrule
        CET* & MEL2C & 97.76 $\pm$ 0.07 & 96.13 $\pm$ 0.23 & 95.18 $\pm$ 0.17 & 99.20 $\pm$ 0.80 & 94.14 $\pm$ 0.31 & 90.57 $\pm$ 0.10 \\ \midrule
        Majority Vote & MRL & 97.48 $\pm$ 0.00 & 92.48 $\pm$ 0.05 & 62.55 $\pm$ 0.44 & 99.58 $\pm$ 0.00 & 94.13 $\pm$ 0.00 & 94.14 $\pm$ 0.00  \\
        UnionNet & MRL & 93.34 $\pm$ 0.45 & 95.59 $\pm$ 0.11 & 93.54 $\pm$ 0.21 & 74.08 $\pm$ 0.92 & 85.94 $\pm$ 0.58 & 87.50 $\pm$ 0.52 \\
        CROWDLAB & MRL & 97.72 $\pm$ 0.02 & 92.40 $\pm$ 0.18 & 61.95 $\pm$ 0.31 & 99.58 $\pm$ 0.00 & 94.09 $\pm$ 0.00 & 92.61 $\pm$ 0.00 \\ \midrule
        \rowcolor{Gray!25} LECOMH & MEL2C & \textbf{98.82 $\pm$ 0.05} & \textbf{98.87 $\pm$ 0.10} & \textbf{96.08 $\pm$ 0.02} & \textbf{99.79 $\pm$ 0.21} & \textbf{94.85 $\pm$ 0.15} & \textbf{94.30 $\pm$ 0.09} \\ \bottomrule
    \end{tabular}}
    \caption{Quantitative comparison of the best accuracy of the SOTA SEL2D~\citep{whoshould_mozannar23}, ME\{L2D,L2C\}~\citep{ijcai2022-344,multil2d} and MRL~\citep{goh2022CROWDLAB, wei2022deep} methods on the human-AI collabotative classification datasets. The SEL2D methods are always pre-trained with LNL techniques, with the single user being simulated with random selection from the pool of three annotators. 
    The ME\{L2D,L2C\} methods are also pre-trained with LNL techniques. 
    The notation * means the results are under zero coverage (i.e., using the annotations from all human experts). 
    The best result per benchmark is marked in bold.}
\end{table*}

\begin{figure*}[tbp]
\centering
\subfloat[Test accuracy vs $\lambda$.]{\label{fig:lambda_acc}\includegraphics[width=.475\linewidth]{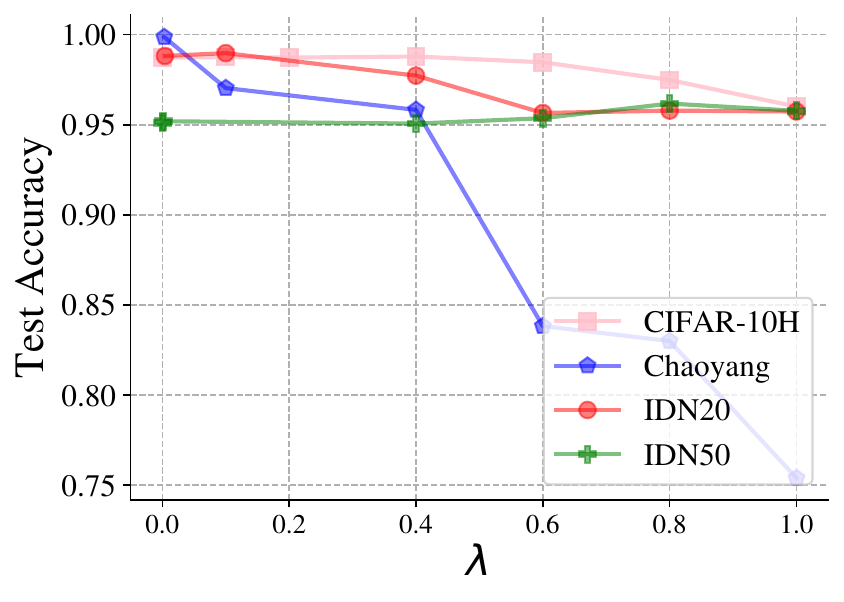}}\hfill
\subfloat[Coverage vs $\lambda$.]{\label{fig:lambda_cost}\includegraphics[width=.475\linewidth]{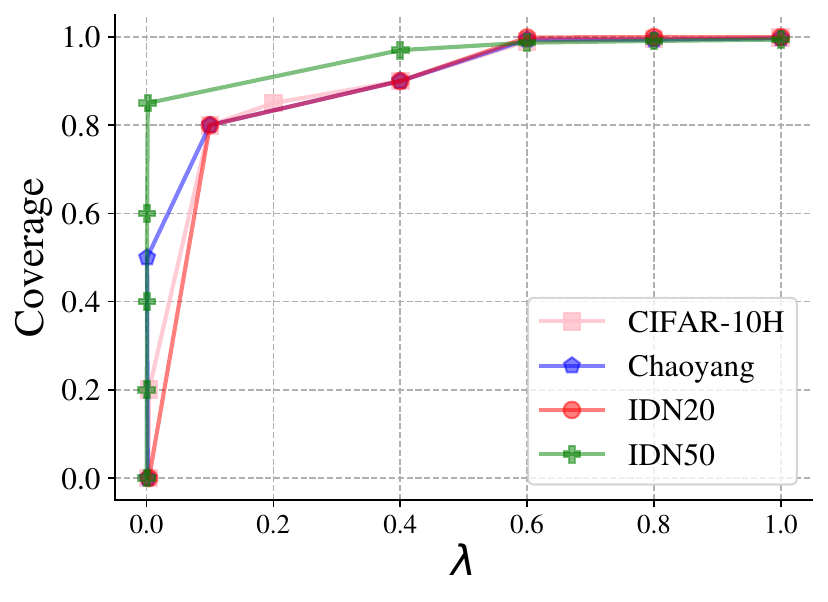}}\\
\caption{Test accuracy vs coverage as a function of $\lambda$ in \cref{eq:loss_function} that weights the collaboration cost in our optimisation.}
\label{fig:lambda}
\end{figure*}

\begin{figure*}[tbp]
\centering
\subfloat[CIFAR-10H.]{\label{fig:cifar10h_n_usrs}\includegraphics[width=.475\linewidth]{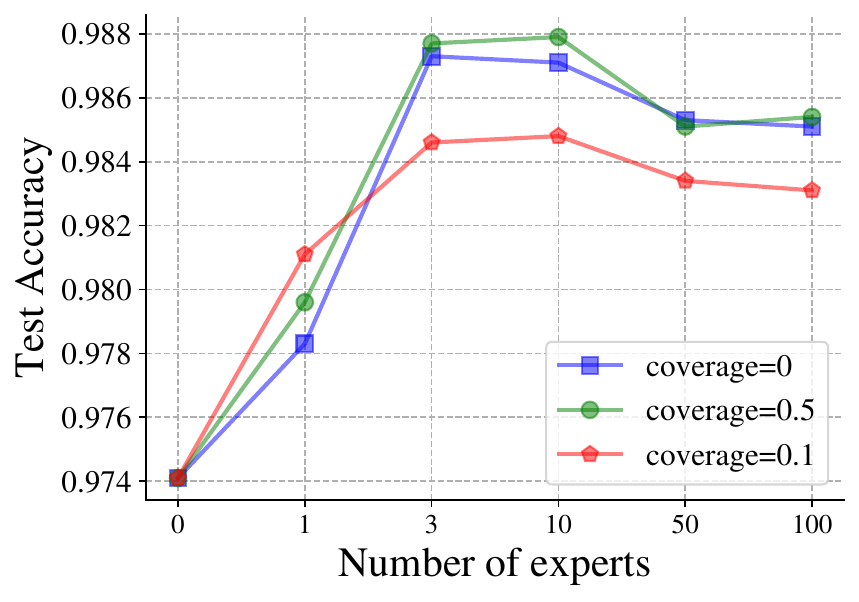}}\hfill
\subfloat[IDN20.]{\label{fig:idn20_n_usrs}\includegraphics[width=.475\linewidth]{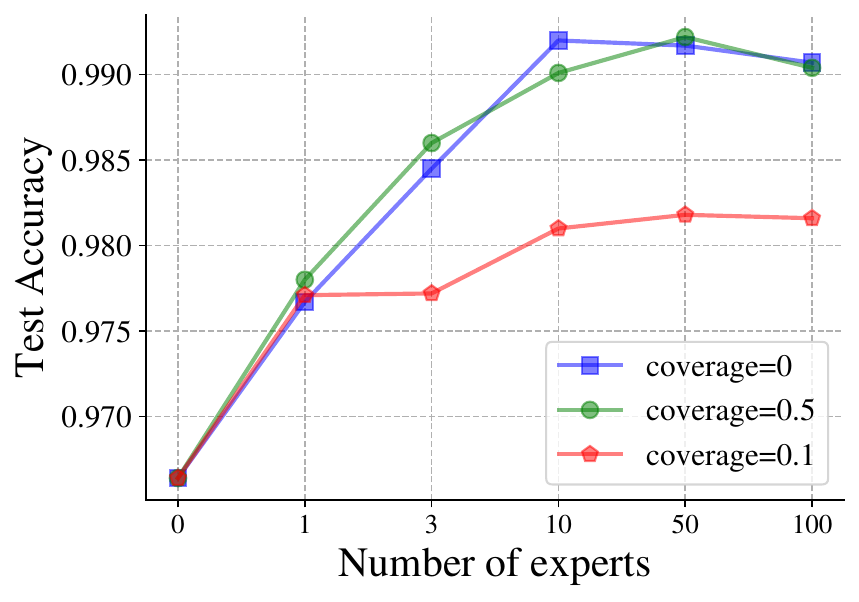}}\\
\caption{Test accuracy vs number of experts at different coverage.}
\label{fig:numofusers}
\end{figure*}

\subsection{Results}


\cref{fig:exp} showcases the accuracy of LECOMH and 
baseline 
methods as a function of coverage in all benchmarks. 
In these graphs, the full coverage point (i.e., coverage equals to 1, meaning only the AI model makes prediction) represents the performance of the LNL pre-trained methods, possibly fine-tuned by the HAI-CC method (resulting in slight variations from the original LNL accuracy). 
Moreover, the results with minimum coverage (i.e., coverage equals to 0) represent the performance of experts alone.

Despite using the same backbone models trained with leading LNL methods and obtaining consensus labels from CROWDLAB, the observed differences highlight the effectiveness of our proposed LECOMH in human-AI collaborative classification techniques in comparison with all baseline methods. 
More specifically, the 
results in~\cref{fig:exp} highlight LECOMH's ability not only to minimise the impact of expert prediction errors, but also to integrate expert information.
In fact, compared to all baselines, LECOMH is the only method that consistently achieves high human-AI collaborative classification accuracy across different levels of coverage, surpassing human and AI accuracies in all benchmarks.

A more fine-grained analysis of LECOMH's results in~\cref{fig:exp} suggest that: 1) for scenarios where human experts have relatively higher or similar accuracy as LNL methods (e.g., CIFAR-10H, Chaoyang, NIH-Aispace opacity and NIH-Nodule/mass), HAI-CC methods can offer significant gains; but 2) for scenarios where users have low accuracy (e.g., IDN20 and IDN50) and LNL methods provide highly accurate predictions, HAI-CC methods offer little improvements over the LNL classifier results.
Hence, as already concluded in~\citep{rastogi2023taxonomy}, it is important to study the conditions that enable HAI-CC to thrive.
Nevertheless, it is interesting to note that for LECOMH, if we set coverage at 50\%, we are always guaranteed to obtain the best possible HAI-CC performance that is usually better (or at least comparable, as in IDN50) than AI or user alone.


In addition to the accuracy versus coverage graphs of~\cref{fig:exp}, we also provide quantitative analyses in~\cref{tab:quantitive_comparison,tab:best_acc_comparison}. 
In particular, the results in~\cref{tab:quantitive_comparison} are the classification accuracy at a fixed coverage=50\%, while the ones in \cref{tab:best_acc_comparison} show the best possible accuracy.
The notation * denotes the results of those methods are under the condition of zero coverage, which means that the prediction depends on the manual annotations from all human experts.
Note that LECOMH surpasses all SE- and ME-HAI-CC methods. LECOMH also outperforms MRL, AI, and human annotators without collaborating with all of the human experts, especially in NIH-AO and NIH-NM datasets. Notably, in CIFAR-10H, IDN20 and IDN50 benchmarks, the LNL AI model has higher accuracy than humans, while in Chaoyang, NIH-AO and NIH-NM datasets, the LNL AI model has lower accuracy than pathologists.

\cref{tab:case,tab:case_coverage5000} present human-AI classification results at the coverage rates of approximately 50\% on the test set of Chaoyang and CIFAR-10H, respectively.
Cases include the test image, human-provided labels ($\mathcal{M}$), LNL AI model prediction ($f_{\theta}(.)$), prediction probability vector by the Human-AI Selection Module ($g_{\phi}(.)$), final prediction by the Collaboration Module ($h_{\psi}(.)$), and ground truth (GT) label. Notably, when the AI model or the users make individual mistakes, the final prediction tends to be correct, highlighting system robustness. $g_{\phi}(.)$ often assigns high probability to the AI model when the AI model appears to be correct, 
indicating reliance on AI predictions.
On the other hand, when the AI model 
seems to be incorrect, 
then $g_{\phi}(.)$ often assigns low probability to the AI model, and high probabilities to the human-AI collaborative classifications, suggesting a reliance on users.

\begin{table*}[t]
\centering
\scalebox{0.75}{
\begin{tabular}{l c c c c c}
\toprule
\bfseries Image & $\mathcal{M}$ & $f_{\theta}(.)$ & $g_{\phi}(.)$ & $h_{\psi}(.)$ & \bfseries GT \\ \midrule
\begin{minipage}[b]{0.1\columnwidth}
	\centering
	\raisebox{-.5\height}{\includegraphics[width=\linewidth]{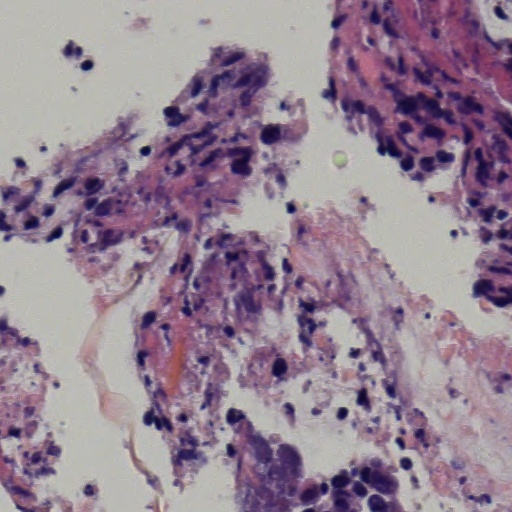}
 }
\end{minipage} 
& adenocarcinoma, normal, adenocarcinoma & normal & {[}0.48, 0.04, 0.03, 0.45{]} & adenocarcinoma & adenocarcinoma  \\ [3ex]
\begin{minipage}[b]{0.1\columnwidth}
	\centering
	\raisebox{-.5\height}{\includegraphics[width=\linewidth]{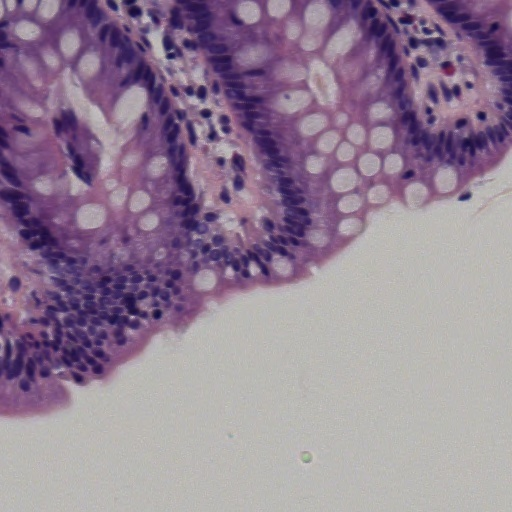}}
\end{minipage} & normal, normal, normal & adenocarcinoma & {[}0.45, 0.06, 0.07, 0.42{]} & normal & normal  \\ [3ex]
\begin{minipage}[b]{0.1\columnwidth}
	\centering
	\raisebox{-.5\height}{\includegraphics[width=\linewidth]{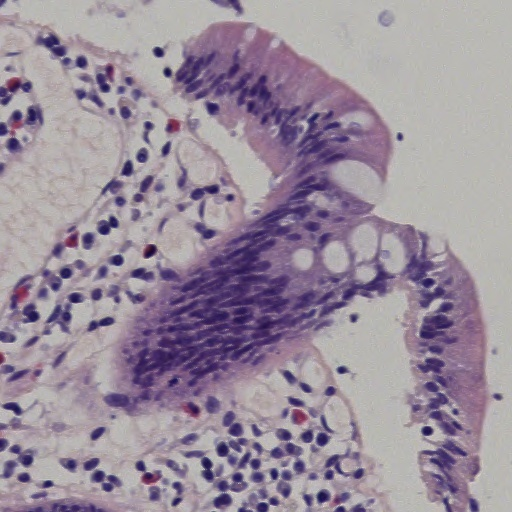}}
\end{minipage} & adenoma, adenoma, adenoma & adenocarcinoma & {[}0.49, 0.07, 0.12, 0.33{]} & adenoma & adenoma  \\ [3ex]
\begin{minipage}[b]{0.1\columnwidth}
 	\centering
 	\raisebox{-.5\height}{\includegraphics[width=\linewidth]{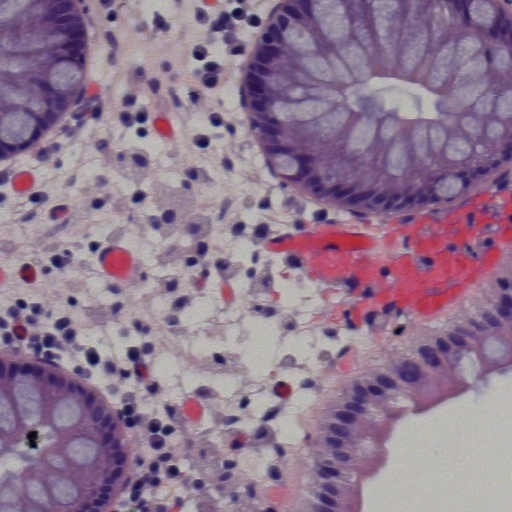}}
 \end{minipage} & serrated, normal, serrated & serrated & {[}0.82, 0.04, 0.03, 0.10{]} & serrated & serrated  \\ [3ex]
\begin{minipage}[b]{0.1\columnwidth}
	\centering
	\raisebox{-.5\height}{\includegraphics[width=\linewidth]{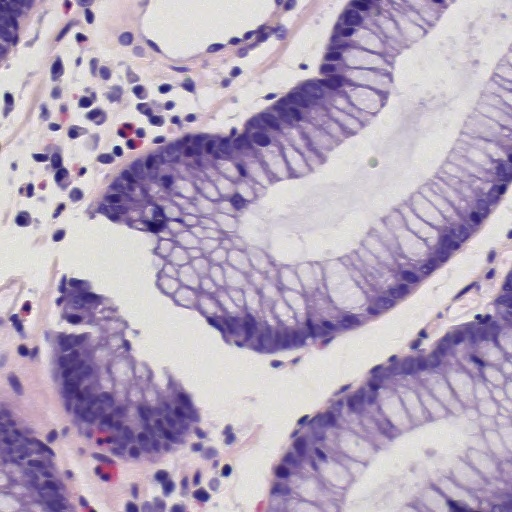}}
\end{minipage} & adenoma, adenoma, serrated & adenoma & {[}0.82, 0.01, 0.02, 0.14{]} & adenoma & adenoma  \\ [3ex]
\begin{minipage}[b]{0.1\columnwidth}
	\centering
	\raisebox{-.5\height}{\includegraphics[width=\linewidth]{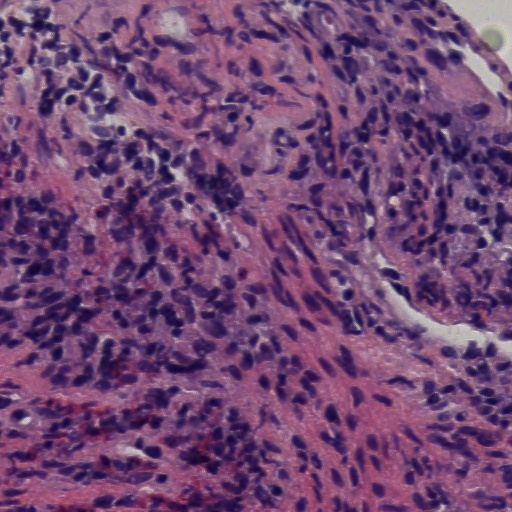}}
\end{minipage} & normal, adenocarcinoma, adenocarcinoma & adenocarcinoma & {[}0.87, 0.03, 0.03, 0.07{]} & adenocarcinoma & adenocarcinoma  \\ 
\bottomrule 
\end{tabular}}
\caption{Human-AI classification (coverage $\approx 50\%$) of Chaoyang test samples, where $\mathcal{M}$ denotes human labels, $f_{\theta}(.)$ is the LNL AI model's classification, $g_{\phi}(.)$ represents the Human-AI Selection prediction probability vector for [AI prediction (1st value), AI + 1 User (2nd value), AI + 2 Users (3rd value), AI + 3 Users (4th value)], $h_{\psi}(.)$ is the final prediction from the Collaboration Module, and GT denotes the ground truth label. }
\label{tab:case}
\end{table*}

\begin{table}[t]
\centering
\begin{tabular}{l@{\hskip 0.75em} c@{\hskip 0.75em} c@{\hskip 0.75em} c@{\hskip 0.75em} c@{\hskip 0.75em} c}
\toprule
\bfseries Image & $\mathcal{M}$ & $f_{\theta}(.)$ & $g_{\phi}(.)$ & $h_{\psi}(.)$ & \bfseries GT \\
\midrule
\begin{minipage}[b]{0.1\columnwidth}
	\centering
	\raisebox{-.5\height}{\includegraphics[width=\linewidth]{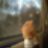}}
\end{minipage} & cat, cat, cat & dog & {[}0.57, 0.05, 0.07, 0.31{]} & cat & cat  \\ [3ex]
\begin{minipage}[b]{0.1\columnwidth}
 	\centering
 	\raisebox{-.5\height}{\includegraphics[width=\linewidth]{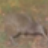}}
 \end{minipage} & bird, bird, bird & frog & {[}0.57, 0.05, 0.04, 0.34{]} & bird & bird \\ [3ex]
\begin{minipage}[b]{0.1\columnwidth}
	\centering
	\raisebox{-.5\height}{\includegraphics[width=\linewidth]{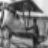}}
\end{minipage} & truck, car, plane & plane & {[}0.70, 0.06, 0.04, 0.20{]} & plane & plane \\ [3ex]
\begin{minipage}[b]{0.1\columnwidth}
	\centering
	\raisebox{-.5\height}{\includegraphics[width=\linewidth]{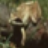}}
\end{minipage} & dog, horse, frog & frog & {[}0.71, 0.06, 0.04, 0.19{]} & frog & frog  \\ [3ex]
\begin{minipage}[b]{0.1\columnwidth}
	\centering
	\raisebox{-.5\height}{\includegraphics[width=\linewidth]{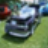}}
\end{minipage} & truck, truck, car & car & {[}0.79, 0.03, 0.05, 0.13{]} & car & car  \\ [3ex]
\begin{minipage}[b]{0.1\columnwidth}
	\centering
	\raisebox{-.5\height}{\includegraphics[width=\linewidth]{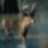}}
\end{minipage} & ship, deer, ship & deer & {[}0.87, 0.02, 0.02, 0.09{]} & deer & deer  \\
\bottomrule 
\end{tabular}
\caption{Human-AI classification (coverage $\approx 50\%$) of CIFAR-10H test samples, where $\mathcal{M}$ denotes human labels, $f_{\theta}(.)$ is the LNL AI model's classification, $g_{\phi}(.)$ represents the Human-AI Selection prediction probability vector for [AI prediction (1st value), AI + 1 User (2nd value), AI + 2 Users (3rd value), AI + 3 Users (4th value)], $h_{\psi}(.)$ is the final prediction from the Collaboration Module, and GT denotes the ground truth label. 
}
\label{tab:case_coverage5000}
\end{table}

\subsection{Ablation Studies}

\begin{table*}[t]
    \centering
    
    \label{tab:ablation}
    \scalebox{.79}{
    \begin{tabular}{l c c c c c c c c c c}
        \toprule
        \bfseries Methods & LNL & MRL & ME & L2C & \bfseries CIFAR-10H & \bfseries IDN20 & \bfseries IDN50 & \bfseries Chaoyang & \bfseries NIH-AO & \bfseries NIH-NM \\
        \midrule
        Human* & & & & & 95.10 & 79.36 & 49.30  & 92.47 & 88.23 & 89.60 \\ \midrule
        LNL & $\surd$ & \textcolor{BrickRed}{\ding{55}} & \textcolor{BrickRed}{\ding{55}} & \textcolor{BrickRed}{\ding{55}} & 97.41 & 96.64 & 95.90  & 82.44 & 86.65 & 87.41 \\
        \midrule
        LECOMH$\ddagger$ & \textcolor{BrickRed}{\ding{55}} & $\surd$ & $\surd$ & $\surd$ & 97.71 (97.95) & 96.94 (96.94) & 67.39 (95.90) & 97.38 (99.58)  & 93.56 (94.65) & 93.67 (94.19)\\
        LECOMH$\diamondsuit$ & $\surd$ & \textcolor{BrickRed}{\ding{55}} & $\surd$ & $\surd$ & 98.22 (98.25) & 97.69 (98.01) & 80.52 (95.90) & 96.83 (100.00) & 92.78 (94.44)  & 92.27 (94.12)\\
        LECOMH$\dagger$ & $\surd$ & $\surd$ & \textcolor{BrickRed}{\ding{55}} & $\surd$ & 97.83 (97.83) & 97.71 (97.80) & 95.79 (95.90) & 92.96 (94.15) & 91.65 (92.61) & 92.46 (93.27)\\
        LECOMH$\S$ & $\surd$ & $\surd$ & $\surd$ & \textcolor{BrickRed}{\ding{55}} & 86.01 (98.17) & 85.93 (93.37) & 54.63 (69.05)  & 69.34 (99.58) & 85.93 (93.37) & 85.32 (93.17) \\
        \midrule
        \rowcolor{Gray!25}LECOMH & $\surd$ & $\surd$ & $\surd$ & $\surd$  & \textbf{98.77 (98.87)} & \textbf{98.82 (98.97)} & \textbf{96.05 (96.10)} & \textbf{99.58 (100.00)} & \textbf{94.63 (95.00)} & \textbf{94.19 (94.39)} \\
        \bottomrule
    \end{tabular}}
    \caption{Accuracy results for the ablation experiments at 50\% coverage (the best accuracy is shown inside brackets). LNL, MRL, ME, L2C denote the utilisation of LNL pre-trained model, the introduction of consensus label via CROWDLAB, the integration of multiple experts,and the cooperation of human experts and AI prediction, respectively. 
    The last row shows the final LECOMH's results.}
\end{table*}

\subsubsection{Study of each LECOMH component}

In this subsection, we analyse the effect of the following LECOMH components: 1) LNL methods to pre-train the the AI model, 2)  multi-rater learning, 3) the role of multiple users for the collaboration, and 4) the integration of the learning to complement module.
\cref{tab:ablation} shows the results of different settings where in each setting, one of the factors is replaced by a baseline approach.

\paragraph{LNL pre-training} By replacing the LNL pre-training by a regular classifier pre-training with early stopping using the noisy labels, we form the LECOMH$\ddagger$. Note in~\cref{tab:ablation} that this model 
performs poorly in scenarios where the LNL AI model surpasses original human labels, such as CIFAR-10H, IDN20 and IDN50. The absence of LNL pre-training is particularly more severe at high label noise rates (e.g., IDN50 with 50\% noise rate).

\paragraph{Multi-rater learning} 
The replacement of MRL approaches, such as CROWDLAB, by simpler methods to obtain consensus labels (e.g., majority voting or sampling a random label as consensus label) forms the approach denoted as LECOMH\(\diamond\). 
As shown in \cref{tab:ablation}, the absence of MRL negatively impacts LECOMH across various cases, with a more pronounced effect at high label noise rates (e.g., IDN50) and evident even in other lower noise rate scenarios when using a random label as consensus (e.g. Chaoyang). 
To further emphasise the value of MRL, we show the accuracy of the consensus label in the training set produced by majority vote, UnionNet~\citep{wei2022deep} and CROWDLAB~\citep{goh2022CROWDLAB} in \cref{tab:consensus}, where results confirm that the MRL approaches, and in particular CROWDLAB, produce more accurate consensus label.

\paragraph{Multiple users for the collaboration} By replacing the collaboration with multiple users with a collaboration with a single user, we form LECOMH$\dagger$. This new model highlights the crucial role of multiple users, particularly in low-noise rate scenarios (e.g., CIFAR-10H, Chaoyang, IDN20, NIH-AO and NIH-NM). In these cases, LECOMH with three users consistently outperforms LECOMH with a single user. However, for IDN50, the reliance on multiple users does not significantly affect LECOMH, indicating that, for high noise rates, the choice between the collaboration with single or multiple users does not have a significant impact.


\paragraph{Integration of the learning to complement module} 
By replacing the collaboration module by a simpler majority voting module, we build the model LECOMH$\S$. 
As shown in \cref{tab:ablation}, the performance of LECOMH$\S$ is reduced significantly compared to the ones with the collaboration module, highlighting the importance of the collaboration module when designing an human-AI system.


\subsubsection{Cost, Scalability and Training Time}

\paragraph{Varying the weight of cost hyperparameter $\lambda$ in \cref{eq:loss_function}} To study the contribution of the cost hyperparameter to the performance of LECOMH, we run several experiments using different values for \(\lambda\) and measuring the performance on several benchmarks. 
The results in \cref{fig:lambda} show the tradeoff between coverage and accuracy when varying \(\lambda\). Smaller values for \(\lambda\) imply higher accuracy results, as shown in \cref{fig:lambda_acc}, but it also leads to smaller coverage (see \cref{fig:lambda_cost}), which is expected due to the lower cost from the querying of human experts. On the other hand, higher values for \(\lambda\) show lower accuracy and larger coverage.


\begin{table}[t]
    \centering
    
    \scalebox{.9}{
    \begin{tabular}{l@{\hskip 0.25em} c@{\hskip 0.75em} c@{\hskip 0.75em} c@{\hskip 0.75em} c@{\hskip 0.75em} c@{\hskip 0.75em} c}
        \toprule
        & \bfseries CIFAR-10N & \bfseries IDN20 & \bfseries IDN50 & \bfseries Chaoyang & \bfseries NIH-AO & \bfseries NIH-NM \\
        \midrule
        Majority Vote & 0.91 & 0.94 & 0.69 & 0.99 & 0.94 & 0.94 \\
        UnionNet & 0.92 & 0.94 & 0.90 & 0.99 & 0.86 & 0.87 \\
        CROWDLAB & 0.98 & 0.99 & 0.98 & 0.99 & 0.94 & 0.96 \\
        \bottomrule
    \end{tabular}
    }
    \label{tab:consensus}
    \caption{Accuracy of the consensus label from majority vote, UnionNet~\citep{wei2022deep} and CROWDLAB~\citep{goh2022CROWDLAB} in the training set.}
\end{table}

\paragraph{Scalability with many users}
In \cref{fig:numofusers}, we explore the scalability of LECOMH to a large number of users. Instead of collaborating with a maximum of three users, as demonstrated in previous experiments, we show in \cref{fig:numofusers} the performance of LECOMH, on CIFAR-10H and IDN20, when it collaborates with between 0 and 100 users. 
The training with CIFAR-10N required simulating numerous users beyond the three available, which is achieved by learning user-specific label-transition matrices and synthesising labels from these matrices. Notably, CIFAR-10H's performance is unsurprisingly optimised for three users, reflecting the synthetic nature of redundant users. On the other hand, for IDN20, accuracy peaks at around 10 to 50 users, suggesting a correlation between problem difficulty and the required number of users for effective collaboration.
Also note that the run-time complexity of our optimisation has a linear increase in terms of the number of users.
In practice, the training time increases from 29s (3 users) to 30s (100 users) per epoch for both CIFAR-10H and IDN20. 

\begin{figure}[t]
    \centering
    \begin{tikzpicture}
        \pgfplotstableread[col sep=&, row sep=\\, header=true]{
            method & time\\
            RS & 23\\
            MoE & 12\\
            LCE & 24\\
            OvA & 13\\
            SP & 17\\
            CC & 24\\
            DIFT & 98\\
            CET & 32\\
            Multi\_L2D & 11\\
            LECOMH & 30\\
        } \mytable
        \pgfplotstablegetrowsof{\mytable}
        \pgfmathsetmacro{\NumRows}{\pgfplotsretval-1}  

        \hspace{-0.5em}
        \begin{axis}[
            width = 0.75\linewidth,
            xbar=0pt,
            xmax=100,
            y=1.5em,
            ytick=data,
            yticklabels from table={\mytable}{method},
            yticklabel style = {font=\small, align=left},
            ytick pos=left,
            xticklabel style = {font=\small},
            xlabel={Time per epoch (in second)},
            axis x line*=bottom,
            axis y line*=left,
            scale only axis,
            enlarge x limits=auto,
            enlarge y limits=auto,
            nodes near coords,
            every node near coord/.append style={font=\footnotesize},
            every axis plot/.append style={fill=mpl_blue, draw=none}
        ]
            \addplot[] table [y expr=\NumRows - \coordindex, x=time]{\mytable};
        \end{axis}
    \end{tikzpicture}
    \caption{Training time/epoch of LECOMH and competing methods on CIFAR-10N.}
    \label{tab:traing_time}
\end{figure}
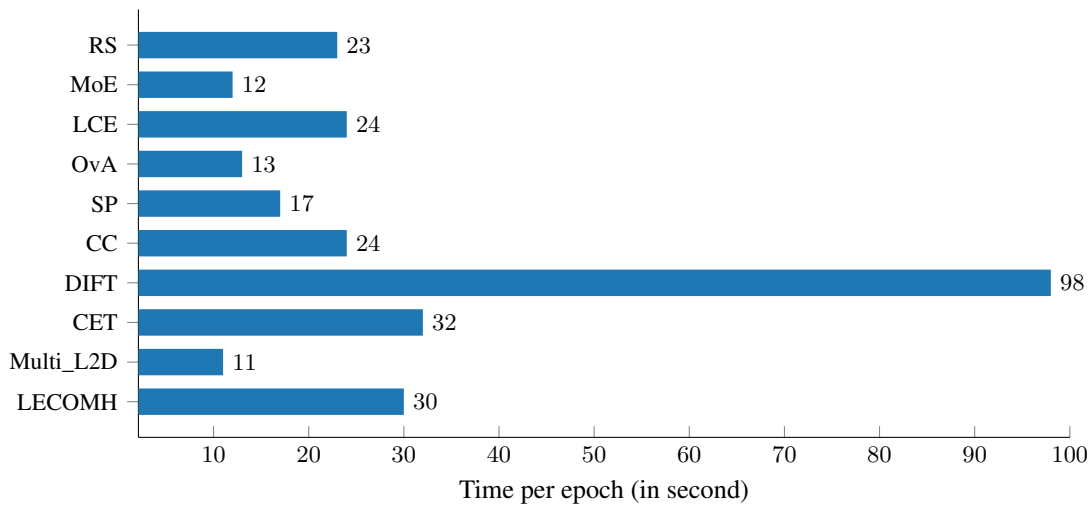

\paragraph{Training time}
In \cref{tab:traing_time}, we compare the training time per epoch of LECOMH and SOTA HAI-CC approaches on CIFAR-10N. Notice that our approach has similar training time compared to other learning to complement approaches, but it has slightly larger training time in comparison to the simpler learning to defer methods.

\section{Conclusion}
\label{sec:conclusion}

We introduced the first human-AI collaborative classification method that
can be trained exclusively from multiple noisy labels to maximise the collaborative classification accuracy of teams of AI and multiple experts, while minimising the collaboration costs, measured by the number of human experts used in HAI-CC.
Additionally, we introduce two new HAI-CC benchmarks that rely on multiple noisy labels in the training and testing sets. Comparative analysis with SOTA HAI-CC methods on our benchmarks demonstrates that LECOMH consistently outperforms the competition, showcasing increased accuracy at comparable collaboration costs. Importantly, LECOMH stands out as the only method enhancing expert labellers and isolated noisy-label learning methods across all benchmarks.

The major limitation of LECOMH is that it is unbiased to any labeller, which implicitly assumes that labellers have similar performance.
Even though this limitation mitigates the complexity involved in the combinatorial selection of specific subsets of labellers, we plan to address this issue 
by exploring a strategy where labellers can be charaterised and selected during training and testing, 
so the system will be able to better adapt to the user's performance. By improving the performance of users who interact with AI systems, we believe that LECOMH has a potential benefit to society given the more accurate outcomes produced by the system and the generally improved performance of labellers.  
Nevertheless, LECOMH may also have potential negative societal impacts.  For example, users may become overconfident on the AI model, which can de-skill some professionals (e.g., radiologists or pathologists). Also, if the training focuses too much on reducing costs, the system will always disregard humans’ predictions, so the training should be performed carefully to achieve a good trade-off between collaboration cost and classification accuracy.

\bibliographystyle{IEEEtranN}
\bibliography{references}

\end{document}